%% file: arxiv.tex
\documentclass{article}

\usepackage{microtype}
\usepackage{graphicx}
\usepackage{subfigure}
\usepackage{booktabs} 
\usepackage{nicefrac}
\usepackage{amsmath,amsfonts,amssymb,mathrsfs}
\usepackage{xcolor}
\usepackage{url}
\newcommand{\kron}{\raisebox{1pt}{\ensuremath{\:\otimes\:}}} 
\newcommand{\kronsum}{\raisebox{1pt}{\ensuremath{\:\oplus\:}}} 

\usepackage{bm}
\newcommand{\mathbold}[1]{\bm{#1}}
\newcommand{\mbf}[1]{\mathbf{#1}}
\newcommand{\vect}[1]{\mbf{#1}}

\newcommand{\T}{\top}    
\newcommand{\dd}{\,\mathrm{d}} 
\newcommand{\R}{\mathbb{R}}    
\newcommand{\N}{\mathrm{N}}   

\newcommand{\KL}[2]{\mathrm{D}_\mathrm{KL}\left[#1\,\|\,#2\right]}

\newcommand{\diff}[2]{\mathrm{\frac{d\mathit{#1}}{d\mathit{#2}}}}
\newcommand{\diffII}[2]{\mathrm{\frac{d^2\mathit{#1}}{d\mathit{#2}^2}}}

\newcommand{\valpha}[0]{\mathbold{\alpha}}

\newcommand{\vmu}[0]{\mathbold{\mu}}

\newcommand{\vtheta}[0]{\mathbold{\theta}}

\renewcommand{\mid}[0]{\,|\,}

\newcommand{\vc}{\mbf{c}}

\newcommand{\vf}{\mbf{f}}

\newcommand{\vh}{\mbf{h}}

\newcommand{\vm}{\mbf{m}}

\newcommand{\vw}{\mbf{w}}
\newcommand{\vx}{\mbf{x}}
\newcommand{\vy}{\mbf{y}}

\newcommand{\MA}{\mbf{A}}

\newcommand{\MF}{\mbf{F}}

\newcommand{\MH}{\mbf{H}}
\newcommand{\MI}{\mbf{I}}

\newcommand{\MK}{\mbf{K}}
\newcommand{\ML}{\mbf{L}}

\newcommand{\MP}{\mbf{P}}
\newcommand{\MQ}{\mbf{Q}}

\newcommand{\MV}{\mbf{V}}
\newcommand{\MW}{\mbf{W}}

\definecolor{mycolor0}{rgb}{0.2667,0.4471,0.7098}
\definecolor{mycolor1}{rgb}{0.1647,0.6706,0.3804}
\definecolor{mycolor2}{rgb}{0.8275,0.2627,0.3059}
\definecolor{mycolor3}{rgb}{0.5216,0.4392,0.7176}
\definecolor{mycolor4}{rgb}{0.8118,0.7255,0.4118}
\definecolor{mycolor5}{rgb}{0.2745,0.7176,0.8157}

\definecolor{mylcolor0}{rgb}{0.6902,0.7686,0.8863}
\definecolor{mylcolor1}{rgb}{0.5451,0.8902,0.6941}
\definecolor{mylcolor2}{rgb}{0.9412,0.7490,0.7647}
\definecolor{mylcolor3}{rgb}{0.8627,0.8392,0.9176}
\definecolor{mylcolor4}{rgb}{0.9569,0.9373,0.8667}
\definecolor{mylcolor5}{rgb}{0.7529,0.9020,0.9373}
\definecolor{mylcolor6}{rgb}{0.8750,0.8750,0.8750}
\definecolor{mygrey}{rgb}{0.25, 0.25, 0.25}

\usepackage{tikz,pgfplots}
\usetikzlibrary{plotmarks,shapes,arrows}
\pgfplotsset{compat=newest} 

\pgfplotsset{/pgf/number format/.cd, 1000 sep={}}
\pgfkeys{/pgf/number format/.cd,1000 sep={}}

\pgfplotsset{every axis/.append style={
		grid style={line width=0.6pt,dotted,gray}}}

\pgfplotsset{every axis/.append style={
		legend style={inner xsep=1pt, inner ysep=0.5pt, nodes={inner sep=1pt, text depth=0.1em},draw=none,fill=none}
}}

\newlength\figureheight
\newlength\figurewidth

\usepackage{hyperref}

\definecolor{cgray}{gray}{0.4}
\newcommand{\comm}[1]{\hfill\textcolor{cgray}{#1}}

\usepackage[accepted]{arxiv}

\renewcommand{\algorithmiccomment}[1]{#1}

\icmltitlerunning{Probabilistic Inference for Nonstationary Audio Analysis}

\begin{document}
	
	\twocolumn[
	\icmltitle{End-to-End Probabilistic Inference for Nonstationary Audio Analysis}
	
	
	
	
	\begin{icmlauthorlist}
		\icmlauthor{William J.\ Wilkinson}{queen-mary}
		\icmlauthor{Michael Riis Andersen}{aalto}
		\icmlauthor{Joshua D.\ Reiss}{queen-mary}
		\icmlauthor{Dan Stowell}{queen-mary}
		\icmlauthor{Arno Solin}{aalto}
	\end{icmlauthorlist}
	
	\icmlaffiliation{aalto}{Department of Computer Science, Aalto University, Finland}
	\icmlaffiliation{queen-mary}{Centre for Digital Music, Queen Mary University of London, United Kingdom}
	
	\icmlcorrespondingauthor{William J.\ Wilkinson}{w.j.wilkinson@qmul.ac.uk}
	
	\icmlkeywords{Gaussian processes, time-frequency analysis, probabilistic models}
	
	\vskip 0.3in
	]
	
	
	
	\printAffiliationsAndNotice{} 
	
	\begin{abstract}
		A typical audio signal processing pipeline includes multiple disjoint analysis stages, including calculation of a time-frequency representation followed by spectrogram-based feature analysis. We show how time-frequency analysis and nonnegative matrix factorisation can be jointly formulated as a spectral mixture Gaussian process model with nonstationary priors over the amplitude variance parameters. Further, we formulate this nonlinear model's state space representation, making it amenable to infinite-horizon Gaussian process regression with approximate inference via expectation propagation, which scales linearly in the number of time steps and quadratically in the state dimensionality. By doing so, we are able to process audio signals with hundreds of thousands of data points. We demonstrate, on various tasks with empirical data, how this inference scheme outperforms more standard techniques that rely on extended Kalman filtering.
	\end{abstract}

	\begin{figure}[!t]
		\centering\scriptsize
		%
		\begin{tikzpicture}
		
		\tikzstyle{box} = [anchor=west]
		
		\node[box] at (0,-0.4) {\includegraphics[width=6cm, height=1.6cm]{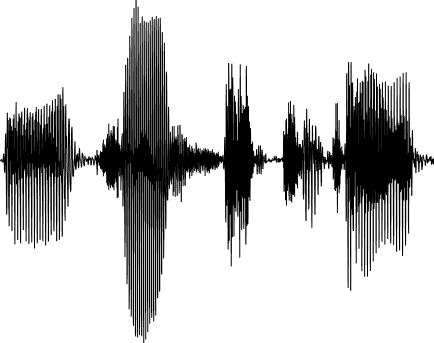}};
		
		\draw[-latex] (0.1,-1.35cm)->(6.2cm,-1.35cm);
		
		\node[box] at (-0.05,1.65cm) {\includegraphics[width=7cm, height=2cm]{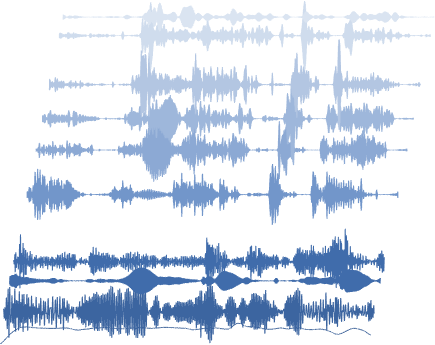}};
		
		\node[box] at (-.25cm,4.1cm) {
			\pgfdeclareimage[width=6cm, height=2cm]{themask}{./figs/teaser/spectrogram_thesis3}
			\pgflowlevel{\pgftransformcm{1}{0}{0.45}{.75}{\pgfpoint{0}{0}}}
			\tikz\node[inner sep=3pt,fill=white,draw=mycolor2,rounded corners=2.5pt, line width=0.35mm, dashed]{\pgfuseimage{themask}};
		};
		
		\node[align=center,text width=5cm,align=center] at (3,-1.6) {Time (sampled at 16 kHz)};
		
		\tikzstyle{label} = [align=center, text width=1.7cm, draw=black!50,rounded corners=2.5pt,rotate=90, inner sep=2pt,minimum height=.7cm, fill=black!10]
		\tikzstyle{math} = [align=center, rotate=90]
		\node[label,text width=1.2cm,draw=mycolor4,fill=mylcolor4] at (-.5cm,-0.4cm) {\bf Audio \mbox{signal $y_k$}};
		\node[math] at (-.5cm,.55cm) {\large $\bm{=}$};
		\node[label,fill=mylcolor0,draw=mycolor0] at (-.5cm,1.75cm) {\bf GP carrier \mbox{subbands $z_d(t)$}};
		\node[math] at (-.5cm,2.9cm) {\large $\bm{\times}$};		
		\node[label,text width=1.4cm,fill=mylcolor2,draw=mycolor2] at (-.5cm,3.9cm) {\bf GP \mbox{spectrogram}};
		
		\node[minimum width=7.56cm, minimum height=1.9cm,draw=mycolor2, line width=0.35mm, dashed,rounded corners=2.5pt,anchor=west] at (-0.8,5.8) {};
		
		\node[box] at (0.4,5.6cm) {\includegraphics[width=6cm,height=1.15cm]{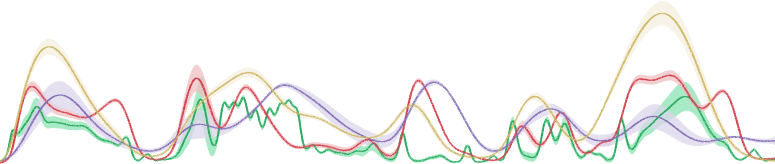}};
		\node[box] at (-.65,5.65cm) {\includegraphics[width=.5cm]{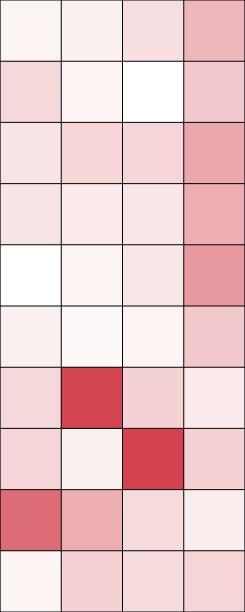}};		
		\node at (0.25,5.65) {\large $\bm{\times}$};
		\node at (3,6.5) {GP spectrogram $=$ NMF weights ($\MW$) $\times$ positive modulator GPs ($g_n(t)$)};
		
		\draw[-latex, draw=mygrey] (6.1,0.6)->(7.15,2.6);
		
		{\fontsize{6.5}{6.5}\selectfont
			\node[align=center,text width=5cm,align=center,rotate=62, text=mygrey] at (6.8,1.5) {Freq. (Hz)};
		}
		
		\end{tikzpicture}
		\vspace{-0.15cm}
		\caption{Graphical introduction to nonstationary modelling of audio data. The input (bottom) is a sound recording of female speech. We seek to decompose the signal into Gaussian process carrier waverforms (\textcolor{mycolor0}{blue block}) multiplied by a spectrogram (\textcolor{mycolor2}{red block}). The spectrogram is learned from the data as a nonnegative matrix of weights times positive modulators (top).}
		\label{fig:teaser}
		\vspace{-0.5cm}
	\end{figure}
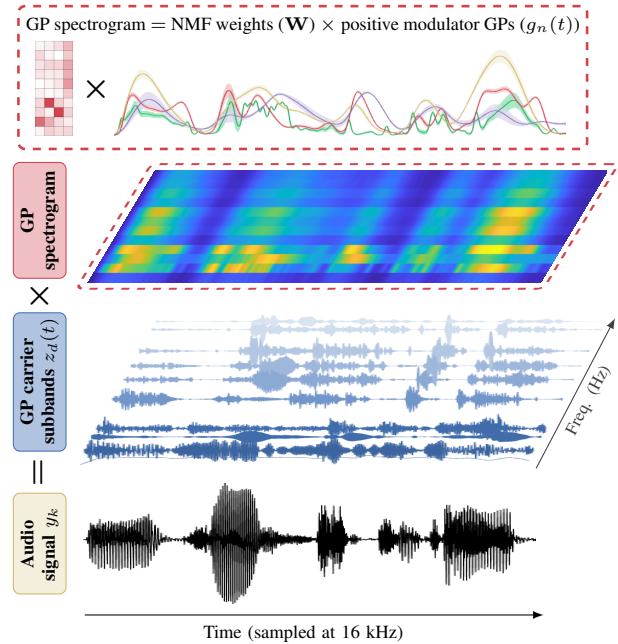

	\renewcommand{\thefootnote}{\fnsymbol{footnote}}
	\section{Introduction}
	\label{sec:introduction}
	Uncovering the high-resolution spectral and temporal information present in a natural auditory scene is a challenging task. Loosely following the approach taken by the human auditory system, we decompose a one-dimensional audio signal into its high-dimensional set of time-varying spectral components, and then utilise the statistical features of these components to perform some auditory task such as classification or source separation. The highly ill-posed nature of this decomposition necessitates the use of prior information about the behaviour of the spectral components, which strongly encourages a probabilistic modelling perspective.
	
	A typical (non-probabilistic) way to perform feature analysis on an audio signal is to apply nonnegative matrix factorisation (NMF) to the amplitude components of a time-frequency (TF) representation -- the spectrogram. As outlined in \cite{turner-time14}, this approach is limited since it discards phase information calculated during the TF stage, as well as dependencies between TF coefficients. It also fails to capture and share any uncertainty information between the analysis stages.
	
	Moreover, the map that takes the waveform to the space of TF coefficients is not a bijection. This means that any function operating on the signal in the TF domain, e.g. noise removal, might push the signal outside the manifold of realisable waveforms \cite{turner2010statistical}. Hence, the modified TF representation must be projected back to the manifold of valid TF representations before the waveform can be re-synthesized (e.g. \citet{griffin1984signal}). This projection might distort the signal and introduce undesirable artefacts.	
	
	These issues have motivated a large body of research on probabilistic models that operate directly on signal waveforms rather than on TF representations. Such models have been shown to outperform their spectrogram-based counterparts on several tasks, including source separation \cite{liutkus2011gaussian, alvarado2018sparse, magron2019complex}, audio inpainting and denoising \cite{badeau2014multichannel, turner-time14}. The limitations of spectrogram analysis have also motivated end-to-end machine learning algorithms for audio generation \cite{pmlr-v70-engel17a, NIPS2018_8023}, generally based on neural networks that require large amounts of training data. In this paper we leverage prior knowledge to construct a probabilistic model that enables inference and learning for short- to medium-duration audio signals.
	
	It has been shown that probabilistic TF analysis can be performed using a Gaussian process (GP) model whose kernel is a sum of quasi-periodic functions \cite{wilkinson2018unifying}. A GP formulation for combining TF analysis with nonnegative matrix factorization (NMF) has also been proposed \cite{turner-time14}. However, the observation mechanism in this joint model is a nonlinear function of the latent components, making inference non-trivial. Previous work relies on a suboptimal inference scheme, where the separate model components are updated independently in an iterative fashion. Moreover, inference in GPs typically scales poorly in the number of time steps, making analysis infeasible for long audio signals. Hence, the full potential of probabilistic models for audio analysis has not yet been realised.
	
	In this work, we propose a probabilistic model and an associated scalable inference algorithm that makes end-to-end audio analysis using GPs possible.
	The contributions of this paper are as follows:
	\begin{itemize}
		\item We construct the state space form of a spectral mixture Gaussian process (GP) with nonstationary NMF priors over the amplitude variance parameters, showing that this model is equivalent to a Gaussian time-frequency NMF model (see Fig.~\ref{fig:teaser} for an overview of the idea).
		\item We design an inference procedure for this nonlinear model based on power expectation propagation in the Kalman smoother setting.
		\item We construct the corresponding infinite-horizon GP \citep{Solin+Hensman+Turner:2018} method for this model, which scales as $\mathcal{O}(M^2T)$ in time and $\mathcal{O}(MT)$ in memory, where $M$ is the dimensionality of the state and $T$ the number of time steps.
		\item We show performance of this approximate inference scheme on various tasks, and compare it to the classical signal processing approach: the iterated extended Kalman filter. By doing so, we demonstrate the flexibility of this generative model.
	\end{itemize}
	
	In Sec.~\ref{sec:GTF} we review the background material and related work on Gaussian process--based models for audio analysis. Sec.~\ref{sec:methods} introduces the proposed model and the associated inference algorithm. Sec.~\ref{sec:experiments} demonstrates performance of the proposed method using a set of audio experiments.

	\section{Gaussian Process Time-Frequency Analysis}
	\label{sec:GTF}
	To specify a probabilistic end-to-end model for the audio processing pipeline, we must replace or remodel the standard processing stages with their probabilistic counterparts. Gaussian processes \citep[GPs, ][]{rasmussen2006gaussian} are a flexible tool for specifying probability distributions over functions, and can be deployed in many such cases.
	
	GP models for time series typically admit the form:
	\begin{subequations}\label{eq:GP}
		\begin{align} 
		f(t) &\sim \mathrm{GP}(0, \kappa(t,t')), \label{eq:GP-prior} \\  
		\vy \mid \vf &\sim\prod_{k=1}^{T} p(y_{k} \mid f(t_{k})), \label{eq:GP-lik}
		\end{align}
	\end{subequations}
	where the one-dimensional input $t$ represents time, Eq.~\eqref{eq:GP-prior} defines the Gaussian process prior and Eq.~\eqref{eq:GP-lik} the likelihood (observation) model. The data $\boldsymbol{\mathcal{D}} = \{(t_k, y_k)\}_{k=1}^T$ consist of input--output pairs and $\kappa(t,t')$ is a covariance function encoding the prior assumptions of the latent (hidden) process $f(t)$. 
	
	Following the typical approach \citep[see, e.g., ][for an overview]{Rasmussen+Nickisch:2010} we seek an approximate posterior of the form:
	\begin{equation}\label{eq:posterior}
	q(\vf \mid \boldsymbol{\mathcal{D}}) = \mathrm{N}(\vf \mid \MK\valpha, (\MK^{-1}+\MV)^{-1}),
	\end{equation}
	where the covariance matrix $K_{i,j} = \kappa(t_i,t_j)$ comes from the prior, $\valpha \in \R^T$, and the likelihood precision matrix $\MV$ is diagonal.
	%
	
	The predictive distribution for a test input $t_*$ with training locations $\bm{t}$ is obtained by integrating the Gaussian latent marginal distribution $\mathrm{N}(f_* \mid \mu_{\mathrm{f},*},\sigma^2_{\mathrm{f},*})$, where $ \mu_{\mathrm{f},*} = \MK(t_*,\bm{t}) \valpha$ and $\sigma^2_{\mathrm{f},*} = \MK(t_*,t_*) - \MK(t_*,\bm{t}) ( \MK + \MV^{-1} )^{-1} \MK(\bm{t},t_*)$, against the likelihood $p(y_* \mid f_*)$ to obtain $p(y_*) = \int p(y_* \mid f_*) \, \mathrm{N}(f_* \mid \mu_{\mathrm{f},*}, \sigma^2_{\mathrm{f},*}) \dd f_*$, the predictive distribution describing the unknown $y_*$.
	
	A probabilistic way of learning the hyperparameters $\vtheta$ of the covariance function and the observation model is by maximising the log marginal likelihood function \citep{rasmussen2006gaussian} (or an approximation of it),
	\begin{equation}
	\log p(\vy\mid\vtheta) = \log \int \mathrm{N}(\vf \mid \mathbf{0}, \MK_{\vtheta} ) \, \prod_k p(y_k \mid f_k, \vtheta) \dd \vf.
	\end{equation}
	\vspace*{-18pt}
	\paragraph{Issues in dealing with the latent functions} Given the well-established GP modelling framework, it may seem surprising that these methods are not widely used in audio modelling. However, the prohibitive computational cubic time-scaling in the number of data renders this naive approach useless for most audio applications where data samples are typically acquired at thousands of samples per second (say, 16~kHz).
	
	Standard approaches for speeding up GP inference, such as inducing input \citep{Quinonero-Candela+Rasmussen:2005,Snelson+Ghahramani:2006,Titsias:2009}, interpolation approaches \citep{Wilson+Nickisch:2015-ICML}, stochastic methods \citep{Hensman+Fusi+Lawrence:2013,Krauth+Bonilla+Cutajar+Filippone:2017}, basis function approximations \citep{Lazaro-Gredilla+Quinonero-Candela+Rasmussen+Figueiras-Vidal:2010,Hensman+Durrande+Solin:2018,Solin+Sarkka:2014-hilbert} scale poorly in long (or potentially unbounded) time series models such as audio analysis. Band-structured or Toeplitz methods \citep{Saatcci:2012} work for data whose sampling is fixed, but would, for example, fail in missing data analysis and only be applicable in batch data scenarios.
	
	Recent advances in combining GP models with efficient signal processing methods have lead to schemes that reformulate the GP prior in terms of a state space model and conduct inference by Kalman filtering in {\em linear} time complexity \citep{Reece+Roberts:2010, Sarkka+Solin+Hartikainen:2013}. If the GP prior exhibits Markov structure, these models are exact and no approximations are needed. Recently, \citet{Nickish+Solin+Grigorievskiy:2018} bridged the gap between the state space and kernel based GP methods, by providing a unifying framework for inference in non-Gaussian likelihoods with established inference schemes like the Laplace approximation, direct KL minimisation, variational Bayes, and single-sweep expectation propagation (EP). We build on these state space methods for linear-time inference for GP audio modelling.
	
	\vspace*{-6pt}
	\paragraph{Probabilistic time-frequency analysis} It has been shown that standard approaches to probabilistic time-frequency analysis are equivalent to Gaussian process regression where the GP kernels are a sum of quasi-periodic components \cite{wilkinson2018unifying}. Such kernels, known as spectral mixtures \cite{wilson2013gaussian}, can be written generally as
	\begin{subequations} 
		\begin{align}
		\kappa_{\mathrm{sm}}(t,t') &= \sum_{d=1}^D\kappa_z^{(d)}(t,t'), \label{SMa}\\
		\kappa_z^{(d)}(t,t') &=\sigma^2_d\cos(\omega_d(t-t'))\kappa_d(t,t'), \label{SMb} 
		\end{align}
	\end{subequations}
	and $\kappa_d$ is free to be chosen, but is typically from the Mat\'ern class. Parameters $\omega_d$ determine the periodicity of the kernel components, which can be interpreted as the centre frequencies of the filters in a probabilistic filter bank. By choosing the exponential kernel $\kappa_d(t,t')=\exp(|t-t'|/\ell_d)$ we recover exactly the probabilistic phase vocoder \cite{cemgil2005probabilistic}, and the lengthscales $\ell_d$ control the filter bandwidths.
	
	The drawback of this model for audio data is that it assumes independence across frequency channels. Correlation between amplitudes of harmonics or modes of vibration is crucial for representing audio signals and is a key component of auditory perception \cite{turner2010statistical, mcdermott2013summary}. This motivates a model that explicitly captures these intra-channel correlations. However, such models no longer observe data through linear combinations of the latent functions, and typical techniques for dealing with these cases tend to fail due to the complex interactions present in audio data. This paper is concerned with addressing these issues.
	
	\vspace*{-6pt}
	\paragraph{Nonnegative matrix factorisation} To capture the desired dependencies across channels, we follow \citet{turner-time14} by utilising nonnegative matrix factorisation (NMF) \cite{lee1999learning}. NMF decomposes a high-dimensional matrix $\mathbf{A} \in \R^{D \times T}$, such as the spectrogram of an audio signal, into a product of two lower-rank nonnegative matrices: a temporal basis $\mathbf{G}$, and a spectral basis $\mathbf{W}$,
	\begin{equation} \label{NMF}
	\mathbf{A} \simeq \mathbf{W}\mathbf{G}.
	\end{equation}
	Typically $\mathbf{W} \in \R^{D \times N}$ and $\mathbf{G} \in \R^{N \times T}$ are learnt by minimising the divergence between the left and right hand sides of Eq.~\eqref{NMF}. In the next section, we place a GP prior over the rows of $\mathbf{G}$ and treat the elements of $\mathbf{W}$ as free parameters of our probabilistic model.		
	
	\section{Methods}
	\label{sec:methods}
	In this section we will first write down the model along with its equivalent presentation as a nonstationary spectral mixture GP. We'll then discuss how it can be constructed as a stochastic differential equation in state space form, before outlining the potential inference methods available.
	
	\subsection{Gaussian Time-Frequency Nonnegative Matrix Factorisation Model (GTF-NMF)}
	\label{sec:model}
	We aim to decompose an input signal $\{y_k\}_{k=1}^T$ into $D$ unknown frequency (oscillator) channels, whose relative amplitudes are modulated by $N$ temporal NMF components. The GP {\em priors} for the $D+N$ latent model component functions are:
	\begin{subequations} 
		\label{eq:prior}
		\begin{align}
		g_n(t) &\sim \mathrm{GP}(0,\kappa^{(n)}_{\mathrm{g}}(t,t')), \quad n = 1,2,\ldots,N, \label{eq:g_n} \\
		z_d(t) &\sim \mathrm{GP}(0,\kappa^{(d)}_{\mathrm{z}}(t,t')), \quad d = 1,2,\ldots,D, \label{eq:z_d}
		\end{align}  
	\end{subequations}
	where $g_n(t)$ denotes the $n^{\textrm{th}}$ temporal NMF component function and $z_d(t)$ the $d^{\textrm{th}}$ frequency channel. The kernel $\kappa_z^{(d)}$ is chosen to be a quasi-periodic function, i.e. the $d^{\textrm{th}}$ component of a spectral mixture, Eq.~\eqref{SMb}. $\kappa_g^{(n)}$ should be determined by our assumptions about the behaviour of the amplitude modulators, such as their smoothness properties.
	
	The {\em likelihood} model is given by:
	\begin{equation} \label{lik}
	y_k = \sum_d a_{d}(t_k)\, z_{d}(t_k) + \sigma_y\,\varepsilon_k,
	\end{equation}
	for square amplitudes (the magnitude spectrogram):
	\begin{equation} \label{amps}
	a_{d}^2(t_k) = \sum_n W_{d,n}\,\psi(g_{n}(t_k)).
	\end{equation} 
	Positivity of the NMF components is enforced by a link function, the softplus $\psi(g_{n}) = \log(1+\mathrm{e}^{g_{n}})$. $\mathbf{W} \in \mathbb{R}^{D \times N}$ are the NMF weights determining which modulators affect which oscillators. If we set $N<D$, then the model captures amplitude behaviour shared across frequency channels. 
	
	Note that if we set $a_{d}(t_k)=1, \, \forall \,d,k$ then Eq.~\eqref{lik} reduces to standard probabilistic time-frequency analysis, the model given in \citet{wilkinson2018unifying}. If we discard $z_{d}(t_k)$ by calculating a fixed spectrogram, such that $a_{d}^2(t_k)$ become our observations, then Eq.~\eqref{amps} is standard temporal NMF \citep{bertin2010enforcing}. Further removing the GP prior over $g_n$ brings us back to the NMF model in Eq.~\eqref{NMF}.
	
	Fig.~\ref{fig:teaser} shows the model diagrammatically -- the frequency channel subbands $z_d$ are $D$ independent, unit variance GPs with quasi-periodic covariance functions. The modulators $g_n$ and the NMF weights constitute a model for the spectrogram, the squared amplitudes of the frequency channels.
	
	The inference methods we will next present allow for any choice of $\kappa_g$, $\kappa_z$, so long as they can be written in state space form, either approximately or exactly. See \citet{Sarkka+Solin:inpress} for a guide to writing kernels in the appropriate way. We focus on the Mat\'ern kernel class due to their strong connection to autoregressive filters, and because their parameters have convenient interpretations for our task -- their lengthscales and variances relate to the bandwidth and scale of the filters in a filter bank \cite{wilkinson2018unifying}.
	
	If we write down our model in its hierarchical form, we observe a striking similarity to the nonstationary spectral mixture GPs presented in \citet{remes2017non}. This hierarchical form has a hyper-GP prior $g_n(t) \sim \mathrm{GP}(0,\kappa^{(n)}_{\mathrm{g}}(t,t'))$ for each component with an NMF-like positivity mapping $\alpha_d^2(t) = \sum_n W_{d,n}\,\psi(g_n(t))$, and the final model becomes:
	\begin{subequations} 
		\begin{align}
		z(t) &\sim  \mathrm{GP} \bigg(0, \sum_{d=1}^D \alpha_d(t)\alpha_d(t')\cos(\omega_d(t-t'))\kappa_d(t,t')\bigg),\label{nonSMc} \\
		y_k &= z(t_k) + \sigma_y\, \varepsilon_k. \label{nonSMd}
		\end{align}
	\end{subequations}
	This is a nonstationary spectral mixture GP with fixed frequencies $\omega_d$ and lengthscales $\ell_d$, with an NMF mapping in the GP prior over the time-varying amplitude variances $\alpha_d^2(t)$. This equivalence means that the inference methods laid out in Secs.~\ref{sec:EKF} and \ref{sec:EP} also apply to nonstationary spectral mixtures, as do their formulation as SDEs in Sec.~\ref{sec:ss}.

	\subsection{State Space Methods for the Latent Functions}
	\label{sec:ss}
	For scalable computation, we transform the GP model in Eq.~\eqref{eq:prior} into state space form by mapping the associated covariance functions to stochastic differential equations (SDEs). If the GP priors admit (high-order) Markovian structure (as they do in our case), the model has an exact representation in terms of an SDE \citep[see][for examples and discussion]{Solin:2016}. In continuous time, the system of independent GP priors is given by the following linear time-invariant SDE:
	\begin{equation}\label{eq:sde}
	\dot{\vx}(t)=\MF\,\vx(t)+\ML\,\vw(t),
	\end{equation}
	where $\MF\in\R^{M\times M}$ and $\ML\in\R^{M\times S}$, for $S=2D+N$, are the feedback and noise effect matrices, respectively. The driving process $\vw(t)\in\mathbb{R}^{S}$ is a multivariate white noise process with spectral density matrix $\MQ_{c}\in\mathbb{R}^{S\times S}$.
	
	The state $\vx(t)$ corresponds to a stacked multi-output stochastic process representing the GP priors $z_d(t), d=1,\ldots,D$ and $g_n(t), n=1,\ldots,N$. Each of the GP components have a representation in terms of submatrices of $\MF$, $\ML$, and $\MQ_\mathrm{c}$.
	
	The SDE representation of the $D+N$ Gaussian process priors can be written in the following block-Kronecker form:
	\begin{subequations}
	\begin{align}
	\MF = \mathrm{blkdiag}(&\MF_\mathrm{cos}^{(1)} \kronsum \MF_\mathrm{mat}^{(1)}, \ldots, \MF_\mathrm{cos}^{(D)} \kronsum \MF_\mathrm{mat}^{(D)}, \nonumber \\
	&\MF_\mathrm{mat}^{(1)}, \ldots, \MF_\mathrm{mat}^{(N)}), \\
	\ML = \mathrm{blkdiag}(&\ML_\mathrm{cos}^{(1)} \kron \ML_\mathrm{mat}^{(1)}, \ldots, \ML_\mathrm{cos}^{(D)} \kron \ML_\mathrm{mat}^{(D)}, \nonumber \\
	&\ML_\mathrm{mat}^{(1)}, \ldots, \ML_\mathrm{mat}^{(N)}), \\
	\MQ_\mathrm{c} = \mathrm{blkdiag}(&\MI_2 \kron \MQ_\mathrm{c,mat}^{(1)}, \ldots, \MI_2 \kron \MQ_\mathrm{c,mat}^{(D)}, \nonumber \\
	&\MQ_\mathrm{c,mat}^{(1)}, \ldots, \MQ_\mathrm{c,mat}^{(N)}),                        
	\end{align}
	\end{subequations}
	where `$\!\kronsum\!$' and `$\!\kron\!$' denote the Kronecker sum and product. The submatrices $\MF_\mathrm{mat}^{(1)}$, $\MF_\mathrm{cos}^{(1)}$, $\ML_\mathrm{mat}^{(1)}$ etc. correspond to the matrices that make up the SDE representation for the Mat\'ern and cosine kernels \cite{solin2014explicit}. Here we have assumed a Mat\'ern kernel for $\kappa_d$, $\kappa_n$, but this can be altered as necessary.
	
	The audio data (observations) are evenly spaced in time, which simplifies the discrete-time solution to the SDE in Eq.~\eqref{eq:sde}. For discrete input values $t_k$, this translates into
	\begin{equation}\label{eq:ss}
	\vx_{k} \sim \mathrm{N}(\MA \, \vx_{k-1}, \MQ)
	\end{equation}
	with $\vx_{0}\sim\mathrm{N}(\mathbf{0},\MP_{0})$. The discrete-time dynamical model is solved through a matrix exponential $\MA = \exp(\MF\,\Delta t)$. For stationary covariance functions, the process noise covariance is given by $\MQ=\MP_{\infty}-\MA\,\MP_{\infty}\,\MA^\T$. The stationary state (corresponding to the initial state $\MP_0$) is distributed by $\vx_{\infty}\sim\mathrm{N}(\mathbf{0},\MP_{\infty})$ and the stationary covariance can be found by solving the Lyapunov equation $\dot{\MP}_{\infty}=\MF\,\MP_{\infty}+\MP_{\infty}\,\MF^\T+\ML\,\MQ_{c}\,\ML^\T=\mathbf{0}$.

	\subsection{Linearisation-Based Inference}
	\label{sec:EKF}
	In classical signal processing, the most widely used technique for dealing with nonlinear/non-Gaussian inference problems in state space models is the {\em extended Kalman filter} \citep[EKF, ][]{Jazwinski:1970,Bar-Shalom+Li+Kirubarajan:2001}. The EKF, together with the backward-pass known as the extended Rauch--Tung--Striebel smoother, provides a means of approximating the state distributions $p(\vect{x} \mid \vect{y}_{1:T})$ with Gaussians (corresponding to the time-marginals of Eq.~\ref{eq:posterior}):
	\begin{equation}
	q(\vect{x}_k \mid \boldsymbol{\mathcal{D}}) \simeq \mathrm{N}(\vect{x}_k \mid \vect{m}_{k}, \vect{P}_{k}).
	\end{equation}
	In the EKF, these approximations are formed by first-order linearisations of the nonlinearities \citep[see][for a detailed presentation of the extended Kalman filtering recursion]{Sarkka:2013}. For GPs, a related local linearisation scheme is known as the Laplace approximation, where the approximation is improved iteratively by mode-seeking. In signal processing, iterative versions of the EKF are known as iterated filters, where the iteration is typically in the inner update loop \citep[local iterated EKF, ][]{Jazwinski:1970,Maybeck:1982}. Outer-loop variants which---similar to the GP Laplace method---seek a global approximation are known as the global iterated EKF \citep{Zhang:1997}.
	
	In Alg.~\ref{alg:ekf} in the supplementary material, we present an outer-loop extended Kalman filtering scheme for Laplace approximation-like inference. The local linearisation is done with respect to the measurement (likelihood) model in Eq.~\eqref{lik} by deriving its closed-form Jacobian $\MH_\vx(\vx)$. We consider this algorithm as the baseline for our experiments.

	\subsection{Expectation Propagation in the GTF-NMF Model}
	\label{sec:EP}
	
	The signal processing community has provided linear-time algorithms for scaling linear state space models to huge, unbounded time series. While scalable, these methods are limited to systems that are well approximated by linear models and they are in general not capable of producing accurate inference in the presence of strong nonlinear dependencies such as in the model presented in Eq.~\eqref{lik}. \citet{Nickish+Solin+Grigorievskiy:2018} proposed to combine the classical methods with modern tools for approximate inference, e.g. variational Bayes and assumed density filtering (ADF), to overcome this issue. We generalise this work by extending the ADF algorithm to expectation propagation and thus combining the best methods from the signal processing and machine learning communities.
	
	Expectation propagation \citep[EP, ][]{Minka:2001} and power expectation propagation are methods for approximating intractable probability distributions using tractable distributions from the exponential family. EP is a generalisation of ADF and works by minimising local Kullback-Leibler (KL) divergences in an iterative fashion. Power EP can be seen as a further generalisation of EP that minimises local $\alpha$-divergences rather than KL divergences \cite{Minka:2005}.
	
	Using power EP, we approximate the intractable likelihood terms as follows:
	\begin{align} \label{eq:lik_approx}
	p(y_k\mid\vect{g}_k, \vect{z}_k) &\approx  q_k(\vect{g}_k, \vect{z}_k),
	\end{align}
	where each site approximation $q_k$ belongs to the exponential family. Specifically, we assume that $q_k$ takes the form
	\begin{align} \label{eq:EPapprox}
	q_k(\vect{g}_k, \vect{z}_k) = \prod_{n} \mathrm{N}(g_{n, k}\mid \nu^g_{n, k}, \tau^g_{n, k}) \prod_d \N(z_{d,k}\mid \nu^z_{d, k}, \tau^z_{d, k}),
	\end{align}
	where $\nu^g_{n, k}$ and $\tau^g_{n, k}$ are the precision-adjusted mean and precision, respectively, for $g_{n, k}$ etc. This choice leads to a joint Gaussian posterior approximation. Rather than simply matching the two distributions in Eq.~\eqref{eq:lik_approx}, the EP algorithm iteratively refines the posterior approximation by  updating each site approximation $q_k$ in the context of the so-called \textit{cavity distribution} $q_{-k}$. The cavity distribution for the $k^{\textrm{th}}$ observation is defined by removing the contribution of the $k^{\textrm{th}}$ site approximation from the posterior approximation $q(\vect{g}_k, \vect{z}_k \mid \boldsymbol{\mathcal{D}})$. That is, 
	\begin{align} \label{eq:cavity_dist}
	q_{-k}(\vect{g}_k, \vect{z}_k) \propto  \frac{q(\vect{g}_k, \vect{z}_k\mid\boldsymbol{\mathcal{D}})}{q_k(\vect{g}_k, \vect{z}_k)^\eta}
	\end{align}
	for $\eta \in \left(0, 1\right]$, where $\eta = 1$ corresponds to regular EP and $\eta < 1$ to power EP. 
	
	The $k^{\textrm{th}}$ site approximation $q_k$ is then updated by minimising the KL-divergence between the \textit{tilted distribution} $\hat{p}_k = \frac{1}{Z_k} p(y_k\mid\vect{g}_k, \vect{z}_k)^{\eta}q_{-k}(\vect{g}_k, \vect{z}_k)$ and the power EP approximation $q_{k}(\vect{g}_k, \vect{z}_k)^{\eta} q_{-k}(\vect{g_k}, \vect{z_k})$ such that
	\begin{align} \label{eq:KL_min}
	q_k^*\left(\vect{g}_k, \vect{z}_k\mid\boldsymbol{\mathcal{D}}\right) = \arg\min\limits_{q_k} \KL{\hat{p}_k}{q_{k}^{\eta} q_{-k}},
	\end{align}
	or equivalently, by matching the moments of the two distributions. The normalisation constant $Z_k$ is given by
	\begin{align}\label{eq:Zk}
	Z_k = \mathbb{E}_{q_{-k}}\left[p(y_k\mid\vect{g}_k, \vect{z}_k)^{\eta}\right].
	\end{align}
	The moments of the tilted distribution can be obtained from the first two partial derivatives of $\log Z_k$ with respect to two sets of cavity mean parameters $\{\mu^g_{n,-k}\}_{n=1}^N$ and $\{\mu^z_{d,-k}\}_{d=1}^D$. For a full derivation of the normalisation constant and its derivatives, see the supplementary material.
	
	The resulting expectations are analytically intractable because the likelihood is a nonlinear function of $\vect{g}_k$ and $\vect{z}_k$. We numerically approximate the $N$-dimensional integrals required to calculate the expectations with $9^{\textrm{th}}$-order sigma-point methods \cite{McNamee1967,kokkala2016a}. However, the number of sigma-points required in this $9^{\textrm{th}}$-order approximation scales poorly with the number of NMF components, $\frac{1}{2}(2N^4-4N^3+22N^2-8N+3)$, which slows down inference for large N.
	
	The proposed algorithm is prone to convergence issues. To prevent EP from oscillating, we use \emph{damped} updates for the site parameters \cite{Minka:2002}. That is, the site parameters are updated as a convex combination of the current parameter values and the new parameters values. Given the large amount of damping required, we generally had to run EP for 20 iterations to reach convergence, more than the 5-10 that is often reported in simpler models.
	
	Standard EP scales cubicly in the number of observations. However, by using the Rauch--Tung--Striebel smoother to approximate the marginal posterior distributions $q(\vect{g}_k, \vect{z}_k \mid \boldsymbol{\mathcal{D}})$ in Eq.~\eqref{eq:cavity_dist}, we can reduce the complexity of the algorithm to be linear in the number of observations. 
	The EP algorithm is summarised in Alg.~\ref{algorithm:EP}.

	\newcommand{\MU}{\mathbf{U}}
	\newcommand{\vsigma}{\bm{\sigma}}
	
	\newcommand{\vnu}{\bm{\nu}}
	\newcommand{\vtau}{\bm{\tau}}
	
	\begin{algorithm}[tb!]
		\caption{EP using Kalman smoothing}
		\label{algorithm:EP}
		\begin{algorithmic}
			\STATE {\bfseries Input:} $\{t_k, y_k\}_{k=1}^T$                    \comm{training inputs and targets} \\
			\quad \qquad $\MA$, $\MQ$, $\MH$, $\MP_0$      \comm{discretised state space model} \\
			$\vtau \leftarrow \bm{0}, \: \vnu \leftarrow \bm{0}$             \comm{likelihood eff.\ precision and location}
			\WHILE[\comm{EP loop}]{not converged}
			\FOR[\comm{forward pass}]{$k=1$ {\bfseries to} $T$}
			\IF{$k==1$}
			\STATE $\vm_k \leftarrow \bm{0};\:\: \MP_k \leftarrow \MP_0$                        \comm{init}
			\ELSE
			\STATE $\vm_k \leftarrow \MA \vm_{k-1};\: 
			\MP_k \leftarrow \MA \MP_{k-1} \MA^{\top}\negmedspace+\negmedspace\MQ$ \comm{predict}
			\ENDIF
			\IF{has label $y_k$}
			\STATE $\vmu \leftarrow \MH \vm_k;\: \MU \leftarrow \MP_k \MH^\T;\:
			\vsigma^2 \leftarrow \text{diag}\left( \MH \MU \right)$
			\IF{first EP iteration}
			\STATE $\vtau_{-k} \leftarrow \vsigma^2; \:\: \vnu_{-k} \leftarrow \vmu$ \comm{cavity} \\
			\STATE set $(\vnu_k,\vtau_{k})$ to minimise the KL div.\ in Eq.~\eqref{eq:KL_min} by calculating $Z_k$ in Eq.~\eqref{eq:Zk} and its gradients		
			\ENDIF
			\STATE $\vc_k \leftarrow \vmu \oslash \vtau_k  - \vnu_k$
			
			\STATE $\MK_k \leftarrow \MU \, \left(\vsigma^2 + \vect{1}\oslash \vtau_k \right)^{-1}$
			\STATE $\MP_k \leftarrow \MP_k - \MK_k \MU^\T$  \comm{variance} \\
			\STATE $\vm_k \leftarrow \vm_k + \MK_k \vc_k$     \comm{mean} \\
			\ENDIF
			\ENDFOR
			
			\FOR[\comm{backward pass}]{$k=T-1$ {\bfseries to} $1$}
			\STATE $\vect{G}_k \leftarrow \MP_{k} \, \MA^\T \, (\MA\,\MP_{k}\,\MA^\T + \MQ)^{-1}$ \comm{gain}
			\STATE $\vect{m}_{k} \leftarrow \vect{m}_{k} + \vect{G}_k \, (\vect{m}_{k+1} - \MA\,\vect{m}_{k})$
			\STATE $\vect{P}_{k} \leftarrow \vect{P}_{k} + \vect{G}_k \, (\vect{P}_{k+1} - \MA\,\vect{P}_{k}\,\MA^\T - \MQ) \, \vect{G}_k^\T$
			
			\STATE $\vmu \leftarrow \MH \vm_k;\:\: \vsigma^2 \leftarrow \text{diag}\left( \MH \MP_k \MH^\T \right) $\comm{latent} \\
			
			\STATE $\vtau_{-k} \leftarrow \vect{1} \oslash \vsigma^2 - \eta \vtau_k; \:\: \vnu_{-k} \leftarrow \vect{\vmu} \oslash \vsigma^2 - \eta \vnu_k$ \comm{cavity}
			\STATE set $(\vnu_k,\vtau_{k})$ to minimise the KL div.\ in Eq.~\eqref{eq:KL_min} by calculating $Z_k$ in Eq.~\eqref{eq:Zk} and its gradients
			
			\ENDFOR   
			\ENDWHILE
			
			\STATE $\text{\bf Return:}~\mathbb{E}[g_n(t_k)] = \vh^{\mathrm{g}}_n\vm_k; \mathbb{V}[g_n(t_k)] = \vh^{\mathrm{g}}_n\MP_k \vh^{\mathrm{g}\T}_n$
			\STATE $\phantom{\text{\bf Return:}}~\mathbb{E}[z_d(t_k)] = \vh^{\mathrm{z}}_d\vm_k; \mathbb{V}[z_d(t_k)] = \vh^{\mathrm{z}}_d\MP_k \vh^{\mathrm{z}\T}_d$
			\STATE $\phantom{\text{\bf Return:}}~\log p(\vy\mid\vtheta) \simeq {\sum \log Z_k }$
			
		\end{algorithmic}
		Notation: $\vect{a} \circ \vect{b}$ and $\vect{a} \oslash \vect{b}$ denote the element-wise multiplication and element-wise divison of the vectors $\vect{a}$ and $\vect{b}$, respectively. $\MH$ is the measurement model with rows $\vh$.
	\end{algorithm}	
	
	\subsection{Infinite-Horizon Gaussian Processes}
	\label{sec:IHGP}
	The inference in Alg.~\ref{algorithm:EP} has linear time complexity, $\mathcal{O}(TM^3)$ (with $M \ll T$), with respect to the number of data points $T$, and state dimensionality $M$. The memory scaling is $\mathcal{O}(TM^2)$ due to the need for storing the state covariances at every time step. However, in the case of audio data $T$ can be tens or hundreds of thousands even for short audio segments. This is mainly problematic with regards to the required memory ($M$ typically in the range of 100--1000). For example, for $M=100$, the required memory is in the range of 1.2~Gb per second of data.
	
	To mitigate the memory bottleneck, we use the infinite-horizon GP (IHGP) framework proposed by \citet{Solin+Hensman+Turner:2018}, where the GP is approximated by finding an associated posterior steady state of the filter for each of the $D+N$ latent functions. This way the propagation of the covariance terms in Alg.~\ref{algorithm:EP} can be simplified, leading to a computational time-scaling of $\mathcal{O}(TM^2)$ and memory scaling $\mathcal{O}(TM)$. \citet{Solin+Hensman+Turner:2018} derived their method to work with ADF, but the EP formulation given in Alg.~\ref{algorithm:EP} directly lends itself to the approach by using the cavity parameters for updating the likelihood variance terms. With these changes, the required memory drops by orders of magnitude to 12.2~Mb per second of data.

	\subsection{Hyperparameter Tuning}
	\label{sec:hypers}
	Model learning is difficult in this setting due to the highly correlated nature of the kernel hyperparameters and the non-identifiability of the NMF mapping. We initialise the parameters via frequency domain fitting with the standard probabilistic TF model, as outlined in \citet{wilkinson2018unifying}, which is fast and gives an accurate estimate of the subband frequencies and lengthscales. We initialise the NMF weights using standard NMF applied to a spectrogram calculated with the subband model. Further tuning is then carried out by direct optimisation of the (log) marginal likelihood, $\log p(\vy\mid\vtheta) $, which is calculated during Kalman smoothing as shown in Alg.~\ref{algorithm:EP}. We leave development of a more robust learning scheme to future work.

	\begin{table}[!tb]
		\caption{Performance measures for each inference scheme. \emph{`sim.'} shows fit to observed data $\vy$ in the simulated data experiment (likelihood noise variance is $\sigma^2_y=10^{-4}$). \emph{`mis.'} shows mean missing data imputation results on a dataset of 10 musical instrument sounds, with segments of 20ms removed. Signal-to-noise ratio (in dB, larger is better) and root mean square error (smaller is better). Based on predictive mean. MP is the matching pursuit baseline.}
		\label{tbl:combined}
		\centering
		{\footnotesize\noindent\scriptsize%
			\setlength{\tabcolsep}{4pt}
			\begin{tabular*}{\columnwidth}{@{\extracolsep{\fill}} lccccccc}
				\toprule
				& EP1 & EP20 & IHGP1 & IHGP20 & EKF1 & EKF20 & MP \\  
				\midrule
				RMSE (sim.) & $0.044$ & $\boldsymbol{0.003}$ & $0.042$ & $0.029$ & $0.124$ & $0.128$ & ---\\
				\midrule
				SNR (mis.) & $7.494$ & $\boldsymbol{8.087}$ & $4.520$ & $4.591$ & $3.716$ & $3.735$ & $5.232$\\
				RMSE (mis.) & $0.590$ & $\boldsymbol{0.551}$ & $0.720$ & $0.716$ & $0.746$ & $0.743$ & $0.761$\\
				\bottomrule
			\end{tabular*}\vspace*{-0em}}
	\end{table}
	
	\section{Experiments}
	\label{sec:experiments}
	In this section we compare the proposed inference methods, showing that fully iterated EP is absolutely necessary for inference in the GTF-NMF model, since the iterated EKF and single-sweep EP approaches fail to uncover the latent functions with sufficient accuracy. Our generative model is extremely flexible, and we demonstrate here how it can be applied to three different real world tasks (and one simulated task) with no adjustment of the model or algorithm: missing data synthesis, denoising and source separation. The GTF-NMF performs on a similar level to application specific algorithms (better in missing data imputation, worse in denoising), whilst being much more general. 

	For ease of comparison, in all the real-world experiments we set $D=16$, $N=3$ and tune the parameters via single-sweep EP (ADF), with $\eta=0.75$ and damping of $0.1$. We use these parameters to directly compare the different inference methods (with the exception of the simulated data experiment where we use the known parameters). We use the exponential and Mat\'ern-$\nicefrac{5}{2}$ kernels for $\kappa_d$ and $\kappa_g$.
	The advantages of the infinite-horizon approach become clear when we consider the source separation problem, in which the mixture signal contains multiple sources (leading to a very high-dimensional state space $M=123$), and is 6 seconds in duration ($T=96{,}000$). 
	
	\vspace*{-6pt}
	\begin{figure}[!t]
		\centering\scriptsize
		\setlength{\figurewidth}{.4\textwidth}
		\setlength{\figureheight}{.4\figurewidth}
		\pgfplotsset{yticklabel style={rotate=90}, ylabel near ticks, clip=true,scale only axis,axis on top,clip marker paths,legend style={row sep=0pt},xlabel near ticks,legend style={fill=white}}
		\subfigure[First NMF component, $g_1(t)$]{
			\input{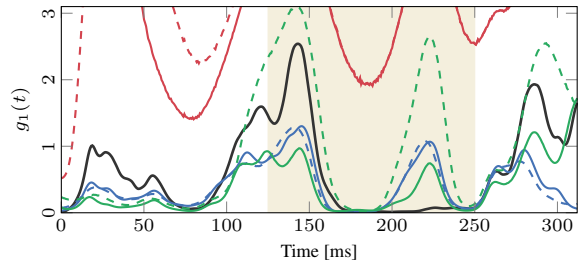}
			\label{fig:mod1}}
		
		\subfigure[Second NMF component, $g_2(t)$]{
			\input{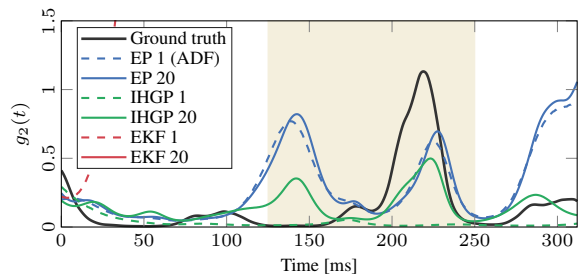}
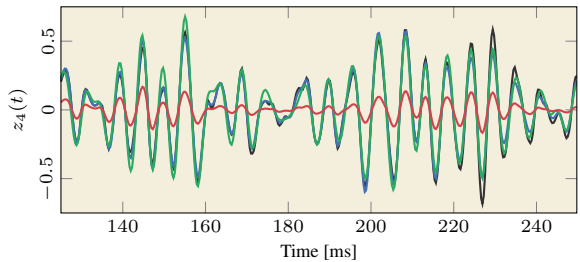
			\label{fig:mod2}}
		
		\subfigure[Short segment of one of the subband signals, $z_4(t)$]{
			\input{./figs/synthetic_data_sub.tex}
			\label{fig:sub}}
		\vspace*{-1em}
		\caption{A simulated data experiment examining the ability of various inference methods to uncover the spectral components $z_d$ and NMF components $g_n$ when the true parameters are known. Due to the ambiguity inherent in the model, (multiple sources of amplitude modulation), uncovering the latents is a difficult task. Standard EP and the IHGP methods far outperform EKF. ``EP 1'' relates to inference with 1 EP iteration (ADF). The iterated methods (dashed lines, each using 20 iterations) resolve the ambiguity better than the single sweep approach, except in the EKF case. Only the mean of the predictive distributions are shown.}
		\label{fig:synthetic_data}
		\vspace*{-0.5em}
	\end{figure}	
	\paragraph{Simulated Data Experiment}
	We set $D=5$, $N=2$ and fix the hyperparameters by hand, before sampling from the generative model to create synthetic data. Fig.~\ref{fig:synthetic_data} shows how each of the proposed inference methods estimates the hidden subband signals and NMF modulators. Uncovering the latents is a highly non-identifiable problem, especially due to the ambiguous nature of the model in which amplitude variation can occur due to variance in the subbands or the modulators. However, EP finds a much better match to the ground truth than EKF, and we see that iterating the IHGP method resolves part of the ambiguity. Table~\ref{tbl:combined} shows how closely the approximate inference methods are able to fit the training data. Since $\sigma^2_y=10^{-4}$, we would hope the RMSE to be below $\sigma_y=0.01$, a feat which only full EP manages.

\begin{figure}[!t]
		\centering\scriptsize
		\setlength{\figurewidth}{.4\textwidth}
		\setlength{\figureheight}{.5\figurewidth}
		\pgfplotsset{yticklabel style={rotate=90}, ylabel near ticks, clip=true,scale only axis,axis on top,clip marker paths,legend style={row sep=0pt},xlabel near ticks,legend style={fill=white}}
		\input{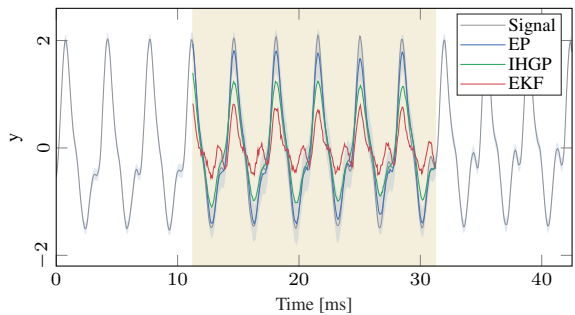}
		\vspace*{-1em}
		\caption{An example of missing data imputation with the GTF-NMF model for each inference method with 20 iterations. Grey signal is the ground truth, a recording of a bamboo flute. The yellow shaded region indicates where the data is missing. Blue shaded area is the $95\%$ confidence region for the EP method.}
		\label{fig:missing_data}
	\end{figure}

	\vspace*{-6pt}
	\paragraph{Missing Data Imputation}
	The generative model handles missing data synthesis naturally by treating the time steps where there are missing data as test locations and making predictions as usual. Table~\ref{tbl:combined} shows the results of the prediction task on a dataset of 10 musical instrument recordings. Fig.~\ref{fig:missing_data} shows an example segment. As a baseline we compare our methods to a well known matching pursuit algorithm \cite{adler2012audio}, which was outperformed by the iterated EP scheme, performing roughly in line with IHGP.

	\vspace*{-6pt}
	\paragraph{Denoising}
	
	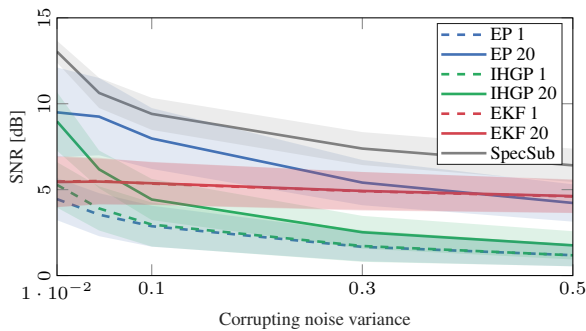
\begin{figure}[!t]
		\centering\scriptsize
		\setlength{\figurewidth}{.4\textwidth}
		\setlength{\figureheight}{.5\figurewidth}
		\pgfplotsset{yticklabel style={rotate=90}, ylabel near ticks, clip=true,scale only axis,axis on top,clip marker paths,legend style={row sep=0pt},xlabel near ticks,legend style={fill=white}}
		\input{./figs/noise_reduction_comparison.tex}
		\vspace*{-1em}
		\caption{Denoising with various inference methods across five levels of corruption noise variance (0.01--0.5). y-axis is the signal-to-noise ratio of the recovered waveform. Mean values across 10 speech signals are shown. Shaded areas are standard error. SpecSub is the spectral subtraction baseline.}
		\label{fig:denoise_comp}
		\vspace{-1em}
	\end{figure}	
	
	Assuming a signal is corrupted by Gaussian noise of known variance, the GTF-NMF model can be adapted to a denoising task by setting the measurement noise variance $\sigma^2_y$ to the appropriate level. Fig.~\ref{fig:noise_reduction} is an example of denoising a speech recording, where the clean signal is corrupted with $\sigma^2_y=0.3$. Fig.~\ref{fig:denoise_comp} shows the denoising results for the various inference methods for five different noise levels. Here we also compare against a spectral subtraction algorithm \cite{ephraim1984speech}. GP models are expected to deal with Gaussian noise well, however the approximate nature of inference in the GTF-NMF prevents it from outperforming  this application-specific approach.
	
\begin{figure}[!t]
		\centering\scriptsize
		\setlength{\figurewidth}{.94\columnwidth}
		\setlength{\figureheight}{.67\figurewidth}
		\pgfplotsset{yticklabel style={rotate=90}, ylabel near ticks, clip=true,scale only axis,axis on top,clip marker paths,legend style={row sep=0pt},xlabel near ticks,legend style={fill=white},axis y line=none,axis x line=bottom,tick align=outside}
		\input{./figs/source_sep.tex}
		\vspace*{-1.1em}
		\caption{Infinite-horizon GP source separation example showing three piano notes (sources) recovered from a mixture signal (top), where two notes are played at a time in the original recording.}
		\label{fig:source_sep}
	\end{figure}
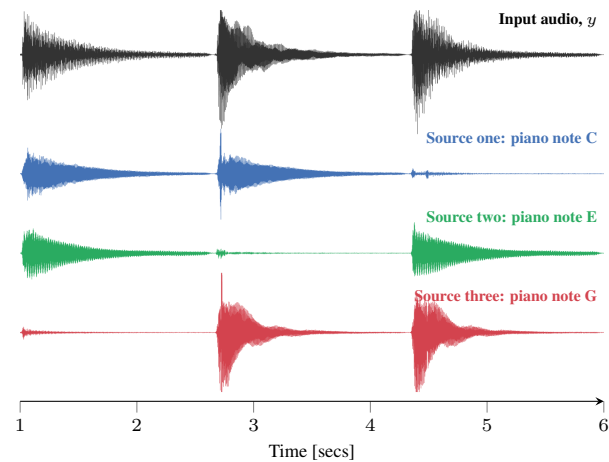

	\vspace*{-6pt}	
	\paragraph{Source Separation}
	As a further demonstration, we follow the approach taken in \citet{alvarado2018sparse} by training the model on musical instrument notes (sources), and then attempting to uncover these sources when they are mixed via summation of their waveforms in a series of two-note chords. The only inference method capable of processing these series of notes is IHGP, due to the computation and memory requirements of stacking the sources in a state space model for 6~seconds of data (sampled at $16$~kHz, $T=96{,}000$, $M=123$). Therefore we cannot compare performance on this task, but we show an example separation result in Fig.~\ref{fig:source_sep}.
	
	\begin{figure}[!t]
		\centering\scriptsize
		\setlength{\figurewidth}{.45\textwidth}
		\setlength{\figureheight}{.28\figurewidth}
		\pgfplotsset{yticklabel style={rotate=90}, ylabel near ticks, clip=true,scale only axis,axis on top,clip marker paths,legend style={row sep=0pt},xlabel near ticks,legend style={fill=white}}
		\input{./figs/noise_reduction_clean.tex}
		\input{./figs/noise_reduction_noisey.tex}		
		\input{./figs/noise_reduction_recon.tex}
		\vspace*{-1.5em}		
		\caption{Spectrograms of a clean, corrupted, and reconstructed signal (from top to bottom) for audio denoising in the GTF-NMF model with inference via EP, applied to a speech signal.}
		\label{fig:noise_reduction}
	\end{figure}
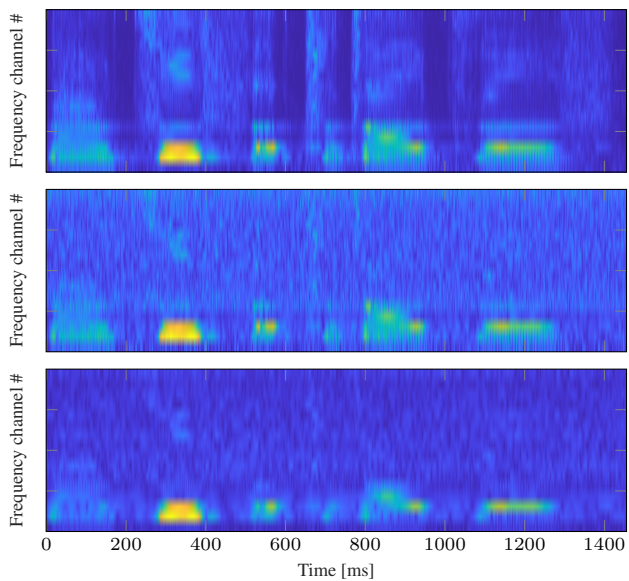
	
	\section{Discussion and Conclusions}
	\label{sec:discussion}

	We have constructed a novel scheme for inference in the Gaussian time-frequency NMF model based on expectation propagation and infinite-horizon GPs, leading to an end-to-end probabilistic approach for audio modelling. By outlining how this model is similar to a nonstationary spectral mixture GP, we have further unified the theory connecting probabilistic machine learning and signal processing.
	
	We demonstrated that our inference scheme consistently outperforms the extended Kalman filtering approach. This suggests that it is indeed necessary to go beyond classical signal processing techniques if we are to build more in-depth nonstationary methods for audio analysis, and that probabilistic modelling has much potential in this domain. By applying it to various real world tasks, we have shown the flexibility of such end-to-end generative models.
	
	For future work, it is necessary to further reduce the inherent computational burden, and to develop more efficient and robust parameter learning schemes to allow these models to become more widely used.
	
	\bibliography{bibliography}
	\bibliographystyle{icml2019}

	
	\clearpage
	\thispagestyle{empty}
	\twocolumn[{\center\baselineskip 18pt
		\toptitlebar{\Large\bf {Supplementary Material for} \\ End-to-End Probabilistic Inference for Nonstationary Audio Analysis}\bottomtitlebar} In this appendix we show further details regarding the globally iterated extended Kalman filter algorithm and the derivations for moment matching in expectation propagation. \vspace*{3em}]
	
	
	\appendix
	
	\section{Global Iterated Extended Kalman Filter}
	Alg.~\ref{alg:ekf} provides pseudo code for the inference algorithm that uses the classical extended Kalman filtering method.
	
	\begin{algorithm}[h]
		\caption{Linearisation-based inference (Laplace approximation scheme) formulated as a global iterated extended Kalman filter.}
		\label{alg:ekf}
		\begin{algorithmic}
			\STATE {\bfseries Input:} $\{t_k\}$, $\vy$, $\MA$, $\MQ$, $\MP_0$ \comm{data and state space model} \\
			\quad \qquad $h(\vx)$, $\MH_\vx(\vx)$             \comm{measurement model and Jacobian}
			$\vm_0 \leftarrow \bm{0}$ \comm{init state mean}
			\WHILE[\comm{iterated EKF loop}]{not converged}
			\FOR[\comm{forward pass}]{$k=1$ {\bfseries to} $T$}
			\IF{$k==1$}
			\STATE $\MP_k \leftarrow \MP_0$ \comm{init state covariance}
			\ELSE
			\STATE $\vm_k \leftarrow \MA\,\vm_{k-1};\: 
			\MP_k \leftarrow \MA\, \MP_{k-1}\, \MA^{\top}{+}\MQ$ \comm{predict}
			\ENDIF
			\IF{has label $y_k$}
			\STATE $v_k \leftarrow {y}_k {-} h(\vect{m}_{k});
			S_k \leftarrow \vect{H}_\vect{x} \, \vect{P}_{k} \, \vect{H}_\vect{x}^\T {+} \sigma_{y}^2$ \comm{inn.}
			\STATE $\vect{k}_k \leftarrow \vect{P}_{k} \, \vect{H}_\vect{x}^\T \, S_k^{-1}$ \comm{gain}
			\STATE $\vect{m}_{k} \leftarrow \vect{m}_{k} + \vect{k}_k \, v_k; 
			\vect{P}_{k} \leftarrow \vect{P}_{k} - \vect{k}_k \, S_k \, \vect{k}_k^\T$
			\ENDIF
			\ENDFOR
			\FOR[\comm{backward pass}]{$k=T-1$ {\bfseries to} $1$}
			\STATE $\vect{G}_k \leftarrow \MP_{k} \, \MA^\T \, (\MA\,\MP_{k}\,\MA^\T + \MQ)^{-1}$ \comm{gain}
			\STATE $\vect{m}_{k} \leftarrow \vect{m}_{k} + \vect{G}_k \, (\vect{m}_{k+1} - \MA\,\vect{m}_{k})$
			\STATE $\vect{P}_{k} \leftarrow \vect{P}_{k} + \vect{G}_k \, (\vect{P}_{k+1} - \MA\,\vect{P}_{k}\,\MA^\T - \MQ) \, \vect{G}_k^\T$
			
			\ENDFOR   
			\ENDWHILE\\
			\comm{measurement row vector $\vh^{\mathrm{g}}_n$ selects $g_n$ from the state}
			\STATE $\text{\bf Return:}~\mathbb{E}[g_n(t_k)] = \vh^{\mathrm{g}}_n\vm_k; \mathbb{V}[g_n(t_k)] = \vh^{\mathrm{g}}_n\MP_k \vh^{\mathrm{g}\T}_n$
			\STATE $\phantom{\text{\bf Return:}}~\mathbb{E}[z_d(t_k)] = \vh^{\mathrm{z}}_d\vm_k; \mathbb{V}[z_d(t_k)] = \vh^{\mathrm{z}}_d\MP_k \vh^{\mathrm{z}\T}_d$
			\STATE $\phantom{\text{\bf Return:}}~\log p(\vy\mid\vtheta) \simeq -\sum_{k=1}^T \frac{1}{2}(\log2\pi S_{k}+v_{k}^{2}/S_{k})$
		\end{algorithmic}
	\end{algorithm}

	\section{Additional Derivations for EP}	
	The normalisation constant required for calculating the posterior in GTF-NMF, as described in Sec.~\ref{sec:EP}, is:
	\begin{gather*}
	\begin{aligned}
	Z_k &= \!&&\mathbb{E}_{q_{-k}}\left[p(y_k\mid\vect{g}_k, \vect{z}_k)^{\eta}\right]\\
	&= && \int...\int \mathrm{N}\Big(y_k \mid \sum_d \sum_n W_{d,n} \psi(g_{n,k}) z_{d,k}, \sigma_y^2 \Big)^\eta \\
	& && \times  \prod_d \mathrm{N}\big(z_{d,k} \mid \mu^z_{d,-k},\zeta^z_{d,-k}\big) \\
	& && \times  \prod_n \mathrm{N}\big(g_{n,k} \mid \mu^g_{n,-k},\zeta^g_{n,-k}\big) \mathrm{d}z_{1,k} ... \mathrm{d}z_{D,k} \mathrm{d}g_{1,k} ... \mathrm{d}g_{N,k} \\
	& = && \mathrm{const}_{\eta}\int...\int \mathrm{N} \Big( y_k \mid \sum_d \sum_n W_{d,n} \psi(g_{n,k}) \mu^z_{d,-k}, \\
	& && \sigma_y^2 + \sum_d \sum_n W_{d,n}^2 \psi(g_{n,k})^2 \zeta^z_{d,-k} \Big) \\
	& && \times  \prod_n \mathrm{N}\big(g_{n,k} \mid \mu^g_{n,-k},\zeta^g_{n,-k}\big) \mathrm{d}g_{1,k} ... \mathrm{d}g_{N,k}
	\end{aligned} \label{eq:appendixZ}
	\end{gather*}	
	for $\mathrm{const}_{\eta}=(2\pi\sigma_y^2)^{\nicefrac{1}{2}(1-\eta)}\eta^{\nicefrac{-1}{2}}$ and where we have used the marginalisation properties of the Gaussian distribution to obtain the last line.
	
	Setting $$m_y=\sum_d \sum_n W_{d,n} \psi(g_{n,k}) \mu^z_{d,-k}$$ and $$v_y=\sigma_y^2 + \sum_d \sum_n W_{d,n}^2 \psi(g_{n,k})^2 \zeta^z_{d,-k}$$ and differentiating w.r.t. $\mu^z, \mu^g$, we get
	
	\begin{gather*}
	\begin{aligned}
	\diff{Z_k}{\mu^z_{d,-k}} &= \!&& \mathrm{const}_{\eta} \int...\int \mathrm{N} \big( y_k \mid m_y, v_y \big) \\
	& &&  \times \sum_n W_{d,n} \psi(g_{n,k})   \frac{y-m_y}{v_y} \\
	& && \times  \prod_n \mathrm{N}\big(g_{n,k} \mid \mu^g_{n,-k},\zeta^g_{n,-k}\big) \mathrm{d}g_{1,k} ... \mathrm{d}g_{N,k},
	\end{aligned} \label{eq:dZdmud}
	\end{gather*}
	
	\begin{gather*}
	\begin{aligned}
	\diff{Z_k}{\mu^g_{n,-k}} &= \!&& \mathrm{const}_{\eta} \int...\int \mathrm{N} \big( y_k \mid m_y, v_y \big) \frac{g_{n,k}-\mu^g_{n,-k}}{\zeta^g_{n,-k}} \\
	& && \times  \prod_n \mathrm{N}\big(g_{n,k} \mid \mu^g_{n,-k},\zeta^g_{n,-k}\big) \mathrm{d}g_{1,k} ... \mathrm{d}g_{N,k},
	\end{aligned} \label{eq:dZdmug}
	\end{gather*}
	
	\begin{gather*}
	\begin{aligned}
	\diffII{Z_k}{\mu^z_{d,-k}} &= \!&& \mathrm{const}_{\eta} \int...\int \mathrm{N} \big( y_k \mid m_y, v_y \big) \\
	& &&  \times \sum_n \left(W_{d,n} \psi(g_{n,k})\right)^2 \left[  \left(\frac{y-m_y}{v_y}\right)^2 - \frac{1}{v_y} \right] \\
	& && \times  \prod_n \mathrm{N}\big(g_{n,k} \mid \mu^g_{n,-k},\zeta^g_{n,-k}\big) \mathrm{d}g_{1,k} ... \mathrm{d}g_{N,k},
	\end{aligned} \label{eq:d2Zdmud2}
	\end{gather*}
	
	\begin{gather*}
	\begin{aligned}
	\diffII{Z_k}{\mu^g_{n,-k}} &= \!&& \mathrm{const}_{\eta} \int...\int \mathrm{N} \big( y_k \mid m_y, v_y \big) \\ 
	& && \times \left[ \left(\frac{g_{n,k}-\mu^g_{n,-k}}{\zeta^g_{n,-k}}\right)^2 - \frac{1}{\zeta^g_{n,-k}} \right] \\
	& && \times  \prod_n \mathrm{N}\big(g_{n,k} \mid \mu^g_{n,-k},\zeta^g_{n,-k}\big) \mathrm{d}g_{1,k} ... \mathrm{d}g_{N,k}.
	\end{aligned} \label{eq:d2Zdmug2}
	\end{gather*}
	
	We can see from the above that all the required integrals are $N$-dimensional, where $N$ is the number of NMF components.
	
	It is now straightforward to obtain the partial derivatives of $\log Z_k$ from the equations above using the chain rule:
	\begin{gather*}
	\begin{aligned}
	\diff{\log Z_k}{\mu^z_{d,-k}} &= \frac{1}{Z_k} \diff{Z_k}{\mu^z_{d,-k}}, \\ 
	\hspace{0.4cm}  \diff{\log Z_k}{\mu^g_{n,-k}} &= \frac{1}{Z_k} \diff{Z_k}{\mu^g_{n,-k}},\\
	\diffII{\log Z_k}{\mu^z_{d,-k}} &= -\frac{1}{Z_k^2} \left(\diff{Z_k}{\mu^z_{d,-k}}\right)^2 + \frac{1}{Z_k} \diffII{Z_k}{\mu^z_{d,-k}}, \\
	\diffII{\log Z_k}{\mu^g_{n,-k}} &= -\frac{1}{Z_k^2} \left(\diff{Z_k}{\mu^g_{n,-k}}\right)^2 + \frac{1}{Z_k} \diffII{Z_k}{\mu^g_{n,-k}}.
	\end{aligned} \label{eq:chain}
	\end{gather*}
	
	We use these derivatives to update the site parameters in Eq.~\eqref{eq:EPapprox}, whilst also converting them back to the precision-adjusted (natural) parameter space, via the following mapping: letting $b_{d,k} = \diff{\log Z_k}{\mu^z_{d,-k}}$ and $c_{d,k}=\diffII{\log Z_k}{\mu^z_{d,-k}}$ and for damping parameter $\rho$,
	\begin{gather*} \label{eq:EP-site-update}
	\begin{aligned}
	\tau^z_{d,k} &= (1 - \rho) \tau^z_{d,k} + \frac{\rho}{\eta} \left( \frac{-c_{d,k}}{1 + \zeta_{d,-k} c_{d,k}} \right) , \\
	\nu^z_{d,k} &= (1 - \rho) \nu^z_{d,k} + \frac{\rho}{\eta} \left( \frac{b_{d,k}-\mu_{d,-k}c_{d,k}}{1 + \zeta_{d,-k} c_{d,k} } \right).
	\end{aligned}
	\end{gather*}
	Mapping to the natural parameter space in this way makes the updates in the EP algorithm more straightforward (see Alg.~\ref{algorithm:EP}). The updates for $\tau^g_{n,k}$ and $\nu^g_{n,k}$ are carried out similarly using the derivatives with respect to $\mu^g_{n,-k}$.

	\vspace{2cm}
	\section{List of Example Audio Samples Used in Experiments}

	\begin{table}[!h]
		\caption{Audio samples used in the experiments. Recordings are mono, $16$kHz sample rate.}
		\label{tbl:audi}
		\centering
		{\footnotesize\noindent%
			\begin{tabular*}{\columnwidth}{@{\extracolsep{\fill}} llc}
				\toprule
				\# & Sample description\hfill & Length [ms] \\  
				\midrule
				01 & speech 1 - male & 723 \\
				02 & speech 2 - male & 595 \\
				03 & speech 3 - male & 941 \\
				04 & speech 4 - male & 854 \\
				05 & speech 5 - male & 514 \\
				06 & speech 6 - female & 769 \\
				07 & speech 7 - female & 731 \\
				08 & speech 8 - female & 959 \\
				09 & speech 9 - female & 1270 \\
				10 & speech 0 - female & 1455 \\
				11 & bamboo flute & 2980 \\
				12 & cello & 3300 \\
				13 & clarinet & 2020 \\
				14 & flute & 2000 \\
				15 & guitar& 2652 \\
				16 & ocarina & 1366 \\
				17 & piano & 1846 \\
				18 & piccolo & 1773 \\
				19 & saxophone & 2296 \\
				20 & toy accordion & 1671 \\
				21 & piano note C & 2000 \\
				22 & piano note E & 2000 \\
				23 & piano note G & 2000 \\
				24 & piano chord mixture & 6000 \\
				25 & piano ground truth C & 6000 \\
				26 & piano ground truth E & 6000 \\
				27 & piano ground truth G & 6000 \\
				
				\bottomrule
		\end{tabular*}}
	\end{table}

\end{document}

%% file: figs/synthetic_data_sub.tex
%
%
\definecolor{mycolor1}{rgb}{0.95686,0.93725,0.86667}%
\definecolor{mycolor2}{rgb}{0.26667,0.44706,0.70980}%
\definecolor{mycolor3}{rgb}{0.16471,0.67059,0.38039}%
\definecolor{mycolor4}{rgb}{0.82745,0.26275,0.30588}%
\begin{tikzpicture}

\begin{axis}[%
xmin=125.0625,
xmax=249.75,
xlabel={Time [ms]},
ymin=-0.75,
ymax=0.75,
ylabel={$z_4(t)$},
axis background/.style={fill=white},
legend style={legend cell align=left,align=left,draw=white!15!black},
width=\figurewidth,
height=\figureheight
]

\addplot[area legend,solid,draw=mycolor1,fill=mycolor1,forget plot]
table[row sep=crcr] {%
	x	y\\
	125.0625	-0.75\\
	250	-0.75\\
	250	0.75\\
	125.0625	0.75\\
	125.0625	-0.75\\
}--cycle;
\addplot [color=white!20!black,solid,line width=0.8pt,forget plot]
table[row sep=crcr]{%
	125.0625	0.205118982252589\\
	125.375	0.23820514190556\\
	125.6875	0.263341637227355\\
	126	0.277687415765383\\
	126.3125	0.286740581162828\\
	126.625	0.267040786152364\\
	126.9375	0.221647615511488\\
	127.25	0.146525176261474\\
	127.5625	0.0472171896219435\\
	127.875	-0.0551336253846379\\
	128.1875	-0.144488330549898\\
	128.5	-0.200303737254766\\
	128.8125	-0.228160496261312\\
	129.125	-0.231098432275842\\
	129.4375	-0.184367156141375\\
	129.75	-0.096205728643435\\
	130.0625	0.00126494441556835\\
	130.375	0.0816186392720853\\
	130.6875	0.132324360632813\\
	131	0.151489342027161\\
	131.3125	0.145873771007378\\
	131.625	0.125118456975176\\
	131.9375	0.0969423536164265\\
	132.25	0.0698031969726955\\
	132.5625	0.0485064246147598\\
	132.875	0.0363147815324181\\
	133.1875	0.03081998961104\\
	133.5	0.0336376307660649\\
	133.8125	0.0385045507973375\\
	134.125	0.0510653064640213\\
	134.4375	0.0486713461269998\\
	134.75	0.0405588315929761\\
	135.0625	0.0274559737731024\\
	135.375	-0.0250241813152343\\
	135.6875	-0.102440391844615\\
	136	-0.188658787257528\\
	136.3125	-0.253198873333423\\
	136.625	-0.285230704522834\\
	136.9375	-0.257373908888171\\
	137.25	-0.19995739381289\\
	137.5625	-0.140806425672918\\
	137.875	-0.0466329548306576\\
	138.1875	0.0607491898464802\\
	138.5	0.161141145488788\\
	138.8125	0.220016112710969\\
	139.125	0.253136123136588\\
	139.4375	0.25093434794873\\
	139.75	0.218249638853857\\
	140.0625	0.156507876617182\\
	140.375	0.0632714267519753\\
	140.6875	-0.0508296641043468\\
	141	-0.169219056998276\\
	141.3125	-0.260935954220284\\
	141.625	-0.302981346597017\\
	141.9375	-0.305255942716482\\
	142.25	-0.279493122839063\\
	142.5625	-0.211712928525595\\
	142.875	-0.132060321565052\\
	143.1875	-0.0400978876422759\\
	143.5	0.0753479461006308\\
	143.8125	0.203882767649943\\
	144.125	0.31671726247188\\
	144.4375	0.402164267300152\\
	144.75	0.452076499257247\\
	145.0625	0.449338219887512\\
	145.375	0.37160050257769\\
	145.6875	0.225585292626693\\
	146	0.0455388720241527\\
	146.3125	-0.140811791315121\\
	146.625	-0.29135309682117\\
	146.9375	-0.372420072206687\\
	147.25	-0.393489123969177\\
	147.5625	-0.346956866329626\\
	147.875	-0.251641658910196\\
	148.1875	-0.142654520828802\\
	148.5	-0.043729610621938\\
	148.8125	0.0421362133158703\\
	149.125	0.100864622084372\\
	149.4375	0.128963603975096\\
	149.75	0.146392032904169\\
	150.0625	0.154361362446781\\
	150.375	0.108593246683086\\
	150.6875	0.0170698287454879\\
	151	-0.0850314356460416\\
	151.3125	-0.193042596886031\\
	151.625	-0.302088157500195\\
	151.9375	-0.381862577070799\\
	152.25	-0.423757715405421\\
	152.5625	-0.399061735202279\\
	152.875	-0.293848583726354\\
	153.1875	-0.153064747967912\\
	153.5	0.00296342234262867\\
	153.8125	0.179296600110416\\
	154.125	0.347522755401095\\
	154.4375	0.463777502584313\\
	154.75	0.539403790674615\\
	155.0625	0.563720622742388\\
	155.375	0.52812718167057\\
	155.6875	0.451790914266204\\
	156	0.33139413104752\\
	156.3125	0.187665597752634\\
	156.625	0.0256065880523742\\
	156.9375	-0.138587468948685\\
	157.25	-0.293121265737003\\
	157.5625	-0.417107091802175\\
	157.875	-0.502424579175887\\
	158.1875	-0.52566349997654\\
	158.5	-0.488011408156543\\
	158.8125	-0.408671712183797\\
	159.125	-0.302530332229535\\
	159.4375	-0.197436252450027\\
	159.75	-0.0987620503247099\\
	160.0625	-0.00960120910385713\\
	160.375	0.0425953901721266\\
	160.6875	0.0635584193310583\\
	161	0.0446264589972912\\
	161.3125	0.0185233753185125\\
	161.625	0.0122434410723323\\
	161.9375	0.0342306316960237\\
	162.25	0.071302082666594\\
	162.5625	0.106954984471873\\
	162.875	0.157129756025654\\
	163.1875	0.181085991372307\\
	163.5	0.195657445443598\\
	163.8125	0.191004016628287\\
	164.125	0.159035168384068\\
	164.4375	0.125569853647786\\
	164.75	0.0633138880303364\\
	165.0625	0.000100696486735206\\
	165.375	-0.0621981341556932\\
	165.6875	-0.117831806129822\\
	166	-0.157076647188849\\
	166.3125	-0.1808948346697\\
	166.625	-0.174189114359136\\
	166.9375	-0.133054809548044\\
	167.25	-0.058502664429671\\
	167.5625	0.0411890750991965\\
	167.875	0.130327778972753\\
	168.1875	0.215951509144206\\
	168.5	0.284085563484893\\
	168.8125	0.300953219558117\\
	169.125	0.250098338625037\\
	169.4375	0.166391407521315\\
	169.75	0.0572241890019465\\
	170.0625	-0.062741155423788\\
	170.375	-0.182532380679626\\
	170.6875	-0.275807466337073\\
	171	-0.317361388017808\\
	171.3125	-0.318042530151187\\
	171.625	-0.30074514872787\\
	171.9375	-0.257131789224878\\
	172.25	-0.179114115523725\\
	172.5625	-0.0836840673739427\\
	172.875	-0.0271000661560671\\
	173.1875	0.00940354617099377\\
	173.5	0.0309373456979625\\
	173.8125	0.0475555514659935\\
	174.125	0.0598060085703837\\
	174.4375	0.0653515006474257\\
	174.75	0.060399354240962\\
	175.0625	0.0458975828605366\\
	175.375	0.0389073933483089\\
	175.6875	0.0470511955306753\\
	176	0.0466027889966151\\
	176.3125	0.04020852122502\\
	176.625	0.00848659348884393\\
	176.9375	-0.0306426893293477\\
	177.25	-0.0690933471199705\\
	177.5625	-0.10259387029351\\
	177.875	-0.113971757049135\\
	178.1875	-0.104389160072246\\
	178.5	-0.0913244780580263\\
	178.8125	-0.0674630097839363\\
	179.125	-0.0491690536184406\\
	179.4375	-0.0465578967125926\\
	179.75	-0.052648160111487\\
	180.0625	-0.0515185607939616\\
	180.375	-0.0453823890853853\\
	180.6875	-0.0337306044907685\\
	181	-0.0084284176522692\\
	181.3125	0.0290858116444131\\
	181.625	0.0773106919712529\\
	181.9375	0.121075328681339\\
	182.25	0.14438991477284\\
	182.5625	0.158949167757859\\
	182.875	0.157364306581382\\
	183.1875	0.144129466455792\\
	183.5	0.14659343831151\\
	183.8125	0.173035296866421\\
	184.125	0.204374515777157\\
	184.4375	0.232734323001256\\
	184.75	0.21810469938071\\
	185.0625	0.161792093944327\\
	185.375	0.0811571012570318\\
	185.6875	-0.0200482870680009\\
	186	-0.124100263451694\\
	186.3125	-0.194144171128433\\
	186.625	-0.232179071920898\\
	186.9375	-0.24336978405832\\
	187.25	-0.234976910093722\\
	187.5625	-0.211087831464275\\
	187.875	-0.163823997891237\\
	188.1875	-0.0960957985809782\\
	188.5	-0.0198498844745692\\
	188.8125	0.0636316310231615\\
	189.125	0.133650933337311\\
	189.4375	0.183961121731503\\
	189.75	0.218748717329501\\
	190.0625	0.211020615524504\\
	190.375	0.160284486861862\\
	190.6875	0.0726829768799624\\
	191	-0.0376723027279573\\
	191.3125	-0.149679623382874\\
	191.625	-0.233635308433705\\
	191.9375	-0.279971854323939\\
	192.25	-0.262807075911031\\
	192.5625	-0.199733502905158\\
	192.875	-0.133551704854359\\
	193.1875	-0.0720084208236128\\
	193.5	-0.0142158538195885\\
	193.8125	0.0279790550953718\\
	194.125	0.0649376464663271\\
	194.4375	0.113648072684977\\
	194.75	0.15934109289886\\
	195.0625	0.209703829867729\\
	195.375	0.255111459551949\\
	195.6875	0.274356078091517\\
	196	0.259416017395254\\
	196.3125	0.215637019201986\\
	196.625	0.128692836235868\\
	196.9375	0.0130129435815458\\
	197.25	-0.110806950404916\\
	197.5625	-0.262190901065221\\
	197.875	-0.404209205751859\\
	198.1875	-0.513341014464652\\
	198.5	-0.556631061586321\\
	198.8125	-0.532167949597912\\
	199.125	-0.470088697156004\\
	199.4375	-0.358821425057014\\
	199.75	-0.203389424048862\\
	200.0625	-0.0521373417960497\\
	200.375	0.102441457840583\\
	200.6875	0.244088414426392\\
	201	0.354416538574136\\
	201.3125	0.430439570368608\\
	201.625	0.488464439742017\\
	201.9375	0.510752072623698\\
	202.25	0.47630735209901\\
	202.5625	0.40101887104867\\
	202.875	0.29339283002786\\
	203.1875	0.188455971451738\\
	203.5	0.0837158303900395\\
	203.8125	-0.0225483462759983\\
	204.125	-0.148614724697647\\
	204.4375	-0.263608754207376\\
	204.75	-0.363982714351335\\
	205.0625	-0.4450187748477\\
	205.375	-0.5037717926972\\
	205.6875	-0.510687534437672\\
	206	-0.462500722531742\\
	206.3125	-0.354535837261101\\
	206.625	-0.203270733682368\\
	206.9375	-0.00967082608087661\\
	207.25	0.182085735349816\\
	207.5625	0.356600100859689\\
	207.875	0.50299821629495\\
	208.1875	0.580008493099989\\
	208.5	0.579464145335918\\
	208.8125	0.516297822117474\\
	209.125	0.411095802960739\\
	209.4375	0.267518855630416\\
	209.75	0.105719966526233\\
	210.0625	-0.0424808109424454\\
	210.375	-0.141261462787311\\
	210.6875	-0.207261453296864\\
	211	-0.228069905895749\\
	211.3125	-0.209343044809651\\
	211.625	-0.141825168714982\\
	211.9375	-0.0349887712655004\\
	212.25	0.101262163623513\\
	212.5625	0.232217022679253\\
	212.875	0.333458327611778\\
	213.1875	0.368056245089781\\
	213.5	0.318524019658106\\
	213.8125	0.204522563200743\\
	214.125	0.050792335314098\\
	214.4375	-0.128083378547933\\
	214.75	-0.296282206464268\\
	215.0625	-0.428987404801982\\
	215.375	-0.507650024973951\\
	215.6875	-0.530381191515029\\
	216	-0.477975759722054\\
	216.3125	-0.372281479189192\\
	216.625	-0.234984902358203\\
	216.9375	-0.0733517961256007\\
	217.25	0.0774867617239907\\
	217.5625	0.208867700632226\\
	217.875	0.306487118383945\\
	218.1875	0.34876615895445\\
	218.5	0.338674431552063\\
	218.8125	0.303855099963629\\
	219.125	0.240612464379099\\
	219.4375	0.146010711819703\\
	219.75	0.0258674717589739\\
	220.0625	-0.105709908288297\\
	220.375	-0.199120739013786\\
	220.6875	-0.285195733671113\\
	221	-0.360439183317324\\
	221.3125	-0.420396460159476\\
	221.625	-0.431325387715416\\
	221.9375	-0.389156077339183\\
	222.25	-0.292508048326998\\
	222.5625	-0.15353806965401\\
	222.875	0.0129873482594668\\
	223.1875	0.176662806495634\\
	223.5	0.315294890777877\\
	223.8125	0.423472189419639\\
	224.125	0.481897360905957\\
	224.4375	0.478398411834878\\
	224.75	0.403262017964141\\
	225.0625	0.281645042696576\\
	225.375	0.09765126718069\\
	225.6875	-0.118523174380737\\
	226	-0.324493032936166\\
	226.3125	-0.515915118672664\\
	226.625	-0.644473533236093\\
	226.9375	-0.682345964755911\\
	227.25	-0.625880892037007\\
	227.5625	-0.473816666837855\\
	227.875	-0.246233902970923\\
	228.1875	-0.0172445335549689\\
	228.5	0.20285943977546\\
	228.8125	0.39041074558864\\
	229.125	0.533493066431737\\
	229.4375	0.583822482476\\
	229.75	0.541503545796955\\
	230.0625	0.429465863782578\\
	230.375	0.261864397941463\\
	230.6875	0.102438172085749\\
	231	-0.0408839285866387\\
	231.3125	-0.181853956129377\\
	231.625	-0.29338610103604\\
	231.9375	-0.345052334884478\\
	232.25	-0.341919168496905\\
	232.5625	-0.309210041561103\\
	232.875	-0.232320362513552\\
	233.1875	-0.125171401395396\\
	233.5	0.00316551107228532\\
	233.8125	0.121788198725246\\
	234.125	0.218263596607754\\
	234.4375	0.292084502336632\\
	234.75	0.309805556662837\\
	235.0625	0.272448248544043\\
	235.375	0.228867112302166\\
	235.6875	0.193706980599394\\
	236	0.156055191774122\\
	236.3125	0.102357491624357\\
	236.625	0.0373785954470358\\
	236.9375	-0.0304946481732514\\
	237.25	-0.101948201177648\\
	237.5625	-0.165451966367401\\
	237.875	-0.204406430038\\
	238.1875	-0.226810799217205\\
	238.5	-0.241716286304366\\
	238.8125	-0.223054303698789\\
	239.125	-0.172217232605773\\
	239.4375	-0.0752480301474528\\
	239.75	0.0263226980912363\\
	240.0625	0.110837258255398\\
	240.375	0.147113390635551\\
	240.6875	0.145807371746111\\
	241	0.111893640116963\\
	241.3125	0.0735355780922366\\
	241.625	0.0492509456370657\\
	241.9375	0.0478924940059293\\
	242.25	0.0473481505390747\\
	242.5625	0.0235406618238814\\
	242.875	-0.0162989160154674\\
	243.1875	-0.0552233665208388\\
	243.5	-0.0818992759248868\\
	243.8125	-0.0961562511013221\\
	244.125	-0.0939947589926424\\
	244.4375	-0.0790354157023732\\
	244.75	-0.055059349393122\\
	245.0625	-0.0301141378459417\\
	245.375	-0.0204972657067887\\
	245.6875	-0.0301719886931374\\
	246	-0.0606021818366188\\
	246.3125	-0.0977278767280416\\
	246.625	-0.137711680828589\\
	246.9375	-0.183693202241217\\
	247.25	-0.205700655657223\\
	247.5625	-0.19454081524266\\
	247.875	-0.157854673897642\\
	248.1875	-0.092939977791862\\
	248.5	0.00331761180605415\\
	248.8125	0.107585327291422\\
	249.125	0.192197160885502\\
	249.4375	0.238236566556228\\
	249.75	0.247434864248098\\
};
\addplot [color=mycolor2,solid,line width=0.8pt,forget plot]
table[row sep=crcr]{%
	125.0625	0.220827803834423\\
	125.375	0.258612339420678\\
	125.6875	0.286395087324367\\
	126	0.300484423753904\\
	126.3125	0.29906877096424\\
	126.625	0.275334924697578\\
	126.9375	0.22253700395135\\
	127.25	0.141169910583363\\
	127.5625	0.0439861322087947\\
	127.875	-0.0550675293778435\\
	128.1875	-0.14261646831486\\
	128.5	-0.207418605533457\\
	128.8125	-0.238384670182434\\
	129.125	-0.226545776563424\\
	129.4375	-0.174427147825145\\
	129.75	-0.0931883208491922\\
	130.0625	-0.00224786061763474\\
	130.375	0.0768002133285513\\
	130.6875	0.127426693094529\\
	131	0.145406858841071\\
	131.3125	0.137739425761737\\
	131.625	0.116148294997988\\
	131.9375	0.0905383898910818\\
	132.25	0.0670679883152432\\
	132.5625	0.04881146161067\\
	132.875	0.0372533183452269\\
	133.1875	0.0320064855590836\\
	133.5	0.0335646107493473\\
	133.8125	0.0408603080656129\\
	134.125	0.0502640430150461\\
	134.4375	0.0535932703359797\\
	134.75	0.0419866742002044\\
	135.0625	0.00906542623362474\\
	135.375	-0.0464350252170728\\
	135.6875	-0.119632218576016\\
	136	-0.194735387964459\\
	136.3125	-0.253965786015008\\
	136.625	-0.282091824008174\\
	136.9375	-0.272826133738355\\
	137.25	-0.228041825040082\\
	137.5625	-0.149426786965239\\
	137.875	-0.046195314592491\\
	138.1875	0.0646386367936258\\
	138.5	0.163903325045046\\
	138.8125	0.235679737853309\\
	139.125	0.273693422335126\\
	139.4375	0.275871298332141\\
	139.75	0.24074709073987\\
	140.0625	0.167676537525359\\
	140.375	0.0621264975536679\\
	140.6875	-0.0612311578535522\\
	141	-0.182565920275045\\
	141.3125	-0.280712814098212\\
	141.625	-0.339619501019013\\
	141.9375	-0.350049274404264\\
	142.25	-0.314444372270646\\
	142.5625	-0.243829677592568\\
	142.875	-0.150107633698089\\
	143.1875	-0.0390692578297097\\
	143.5	0.088492218786761\\
	143.8125	0.22476350466733\\
	144.125	0.353420310149692\\
	144.4375	0.453072807132775\\
	144.75	0.503113532055367\\
	145.0625	0.487934611881485\\
	145.375	0.402421450421108\\
	145.6875	0.253331270666207\\
	146	0.0631657887201402\\
	146.3125	-0.134493154919652\\
	146.625	-0.300789495788912\\
	146.9375	-0.403204102934206\\
	147.25	-0.42714291306709\\
	147.5625	-0.378046051252962\\
	147.875	-0.279634134633103\\
	148.1875	-0.159587215761535\\
	148.5	-0.0441373908569266\\
	148.8125	0.0491770058405602\\
	149.125	0.114666963489164\\
	149.4375	0.153201848636983\\
	149.75	0.164856525967831\\
	150.0625	0.147414340650702\\
	150.375	0.0968839121134538\\
	150.6875	0.0150692307164654\\
	151	-0.0898427583594962\\
	151.3125	-0.203603206150862\\
	151.625	-0.307241251807608\\
	151.9375	-0.380788367515755\\
	152.25	-0.404807236923865\\
	152.5625	-0.370315352833586\\
	152.875	-0.278775090537174\\
	153.1875	-0.144782945563354\\
	153.5	0.0125511870391984\\
	153.8125	0.175012149851569\\
	154.125	0.321784767306295\\
	154.4375	0.435971036436665\\
	154.75	0.504837676384137\\
	155.0625	0.521746909942802\\
	155.375	0.490144960887917\\
	155.6875	0.415106080398354\\
	156	0.306143989933163\\
	156.3125	0.172206170352596\\
	156.625	0.023997174716902\\
	156.9375	-0.126168641392381\\
	157.25	-0.266010659969765\\
	157.5625	-0.380356719836287\\
	157.875	-0.45547716679018\\
	158.1875	-0.480624470661306\\
	158.5	-0.451188131300969\\
	158.8125	-0.378567919126411\\
	159.125	-0.280922662997455\\
	159.4375	-0.178895260696385\\
	159.75	-0.0869327071084939\\
	160.0625	-0.0146677490886233\\
	160.375	0.0310956775978958\\
	160.6875	0.0489078789320485\\
	161	0.0466083533904018\\
	161.3125	0.037755554447225\\
	161.625	0.0368192751346242\\
	161.9375	0.048517532875174\\
	162.25	0.0738035101128605\\
	162.5625	0.108551704985373\\
	162.875	0.146190312593276\\
	163.1875	0.177333893074412\\
	163.5	0.194409319675563\\
	163.8125	0.191086605303559\\
	164.125	0.167905641481865\\
	164.4375	0.127296314504125\\
	164.75	0.0729054824872918\\
	165.0625	0.0101209140195302\\
	165.375	-0.0541279145010235\\
	165.6875	-0.110865534015587\\
	166	-0.153213025125305\\
	166.3125	-0.174510206166068\\
	166.625	-0.165950921202337\\
	166.9375	-0.125362365507947\\
	167.25	-0.0560860396551141\\
	167.5625	0.0323673128418206\\
	167.875	0.122579999872688\\
	168.1875	0.201012711805305\\
	168.5	0.250809400183103\\
	168.8125	0.261775259927952\\
	169.125	0.22929701252489\\
	169.4375	0.157889188096586\\
	169.75	0.0581989340913327\\
	170.0625	-0.0525286513306083\\
	170.375	-0.154806792534819\\
	170.6875	-0.232562252164858\\
	171	-0.2753772803734\\
	171.3125	-0.282631753713842\\
	171.625	-0.25690569345815\\
	171.9375	-0.208124285114771\\
	172.25	-0.147029201301383\\
	172.5625	-0.0849697992327183\\
	172.875	-0.0308515895795714\\
	173.1875	0.0105326181579586\\
	173.5	0.037327859724936\\
	173.8125	0.0536083715598068\\
	174.125	0.0615565985267165\\
	174.4375	0.065155588556637\\
	174.75	0.0644221200572077\\
	175.0625	0.0610702373328511\\
	175.375	0.057528726140392\\
	175.6875	0.0539566288560014\\
	176	0.0471385366789819\\
	176.3125	0.03304207941598\\
	176.625	0.00900803468317591\\
	176.9375	-0.0220711590554245\\
	177.25	-0.052753096815833\\
	177.5625	-0.0787896426716794\\
	177.875	-0.0922167220211413\\
	178.1875	-0.0925149605522679\\
	178.5	-0.0824926679457428\\
	178.8125	-0.0688995832747939\\
	179.125	-0.0573143727040862\\
	179.4375	-0.0529811116530973\\
	179.75	-0.0541696601534939\\
	180.0625	-0.0570737356702405\\
	180.375	-0.0552097715297504\\
	180.6875	-0.044241517817944\\
	181	-0.0200819685502826\\
	181.3125	0.0138147163783349\\
	181.625	0.0527105438591542\\
	181.9375	0.0896896292917061\\
	182.25	0.11748956496276\\
	182.5625	0.134217736329263\\
	182.875	0.145302024938753\\
	183.1875	0.155168657003258\\
	183.5	0.171065290351088\\
	183.8125	0.192517820853906\\
	184.125	0.212252145509241\\
	184.4375	0.218357607194186\\
	184.75	0.19992441171544\\
	185.0625	0.151277884144872\\
	185.375	0.0753047231803149\\
	185.6875	-0.0164249480582101\\
	186	-0.107761686104469\\
	186.3125	-0.18193187582627\\
	186.625	-0.232202312388754\\
	186.9375	-0.251992520126178\\
	187.25	-0.242915913669827\\
	187.5625	-0.210440267861706\\
	187.875	-0.158716264102319\\
	188.1875	-0.0921144692376591\\
	188.5	-0.0144580757943642\\
	188.8125	0.0676036010786265\\
	189.125	0.143275681801995\\
	189.4375	0.197791647400933\\
	189.75	0.220332239681614\\
	190.0625	0.20272283953913\\
	190.375	0.146137372956179\\
	190.6875	0.0562020288628378\\
	191	-0.0504739817690146\\
	191.3125	-0.152666539418788\\
	191.625	-0.227829299624913\\
	191.9375	-0.262039590517078\\
	192.25	-0.253489082998271\\
	192.5625	-0.209035106132633\\
	192.875	-0.146466542833504\\
	193.1875	-0.0798523460403543\\
	193.5	-0.0195402177899021\\
	193.8125	0.0333311163949914\\
	194.125	0.0807275628220447\\
	194.4375	0.129207287334374\\
	194.75	0.180490804124979\\
	195.0625	0.231528001450441\\
	195.375	0.275729176643284\\
	195.6875	0.297808800933228\\
	196	0.289516066156487\\
	196.3125	0.24105862217745\\
	196.625	0.150736776673294\\
	196.9375	0.0233172183322494\\
	197.25	-0.128602285949383\\
	197.5625	-0.287267134463398\\
	197.875	-0.432649657454072\\
	198.1875	-0.542204574163651\\
	198.5	-0.596965581083693\\
	198.8125	-0.585679498447921\\
	199.125	-0.512132127227598\\
	199.4375	-0.389091941237242\\
	199.75	-0.232361014933567\\
	200.0625	-0.0591240927877457\\
	200.375	0.113837631221857\\
	200.6875	0.272982932919822\\
	201	0.405201007864232\\
	201.3125	0.501864573894937\\
	201.625	0.554320292857884\\
	201.9375	0.559073877886134\\
	202.25	0.517312736540627\\
	202.5625	0.43755068879164\\
	202.875	0.332367677904986\\
	203.1875	0.2144558587512\\
	203.5	0.0919193117408\\
	203.8125	-0.0310607391273624\\
	204.125	-0.156514372361716\\
	204.4375	-0.282273517750518\\
	204.75	-0.398731688143235\\
	205.0625	-0.490648147349337\\
	205.375	-0.541156726198189\\
	205.6875	-0.539586841082461\\
	206	-0.482214315317199\\
	206.3125	-0.369033970191131\\
	206.625	-0.207562896583046\\
	206.9375	-0.0150694028594698\\
	207.25	0.18371789198569\\
	207.5625	0.360384895940867\\
	207.875	0.491750918637834\\
	208.1875	0.56336682638985\\
	208.5	0.568751042913889\\
	208.8125	0.510181750855085\\
	209.125	0.400828080755309\\
	209.4375	0.260328397236173\\
	209.75	0.111532292790158\\
	210.0625	-0.0245173880358917\\
	210.375	-0.131910747364966\\
	210.6875	-0.20047314757188\\
	211	-0.224201163563063\\
	211.3125	-0.201729252487132\\
	211.625	-0.133840076423926\\
	211.9375	-0.0283029759869239\\
	212.25	0.0984521053169328\\
	212.5625	0.220036780677211\\
	212.875	0.307520618675163\\
	213.1875	0.337372671848534\\
	213.5	0.299818516627901\\
	213.8125	0.197998096623603\\
	214.125	0.0494615101653123\\
	214.4375	-0.118240144566692\\
	214.75	-0.278492403477973\\
	215.0625	-0.406435743707154\\
	215.375	-0.483481746684591\\
	215.6875	-0.498263142070479\\
	216	-0.450687312609718\\
	216.3125	-0.351755316227844\\
	216.625	-0.22013528453731\\
	216.9375	-0.0736198528973148\\
	217.25	0.0679573992188419\\
	217.5625	0.186911837650887\\
	217.875	0.268669769137655\\
	218.1875	0.307974201750066\\
	218.5	0.308294595320187\\
	218.8125	0.275475029594718\\
	219.125	0.214195309902019\\
	219.4375	0.128964351063927\\
	219.75	0.0281736756384694\\
	220.0625	-0.0742772952150702\\
	220.375	-0.170441787863122\\
	220.6875	-0.255395518564113\\
	221	-0.322588730379225\\
	221.3125	-0.362000941383635\\
	221.625	-0.363248160060528\\
	221.9375	-0.321056544922272\\
	222.25	-0.237568354695946\\
	222.5625	-0.122539951777981\\
	222.875	0.0091533230459176\\
	223.1875	0.139500655456338\\
	223.5	0.253152796516167\\
	223.8125	0.335268656031385\\
	224.125	0.377457139468239\\
	224.4375	0.371780348840928\\
	224.75	0.314733978439245\\
	225.0625	0.211526789639908\\
	225.375	0.0723016824871564\\
	225.6875	-0.0860204420366614\\
	226	-0.243238077405548\\
	226.3125	-0.378490492731084\\
	226.625	-0.467893621077336\\
	226.9375	-0.494662918298598\\
	227.25	-0.451130754201706\\
	227.5625	-0.341955946834127\\
	227.875	-0.186335432119605\\
	228.1875	-0.012501375116089\\
	228.5	0.152840194282077\\
	228.8125	0.287496589590704\\
	229.125	0.374629879424616\\
	229.4375	0.404907405350534\\
	229.75	0.376568630217069\\
	230.0625	0.301448106872512\\
	230.375	0.196697762476864\\
	230.6875	0.0817692927730319\\
	231	-0.0294764448081183\\
	231.3125	-0.126132917654065\\
	231.625	-0.198144261289552\\
	231.9375	-0.237332376719595\\
	232.25	-0.241135248271547\\
	232.5625	-0.212449856596225\\
	232.875	-0.157176653276114\\
	233.1875	-0.0836032714096907\\
	233.5	-0.00121989212770578\\
	233.8125	0.0781817838212254\\
	234.125	0.143501328432305\\
	234.4375	0.184926206587819\\
	234.75	0.199105250802102\\
	235.0625	0.188613592753979\\
	235.375	0.162768164313064\\
	235.6875	0.131492584425501\\
	236	0.0990524399523818\\
	236.3125	0.0626084086403947\\
	236.625	0.0218368039708136\\
	236.9375	-0.0210158500185151\\
	237.25	-0.0633925764262067\\
	237.5625	-0.101778816168233\\
	237.875	-0.135443759030641\\
	238.1875	-0.157535623900002\\
	238.5	-0.161311748566363\\
	238.8125	-0.143066221956524\\
	239.125	-0.104097761094119\\
	239.4375	-0.0485069136021705\\
	239.75	0.0118785362928608\\
	240.0625	0.0620116413197312\\
	240.375	0.091636476193105\\
	240.6875	0.098113038654327\\
	241	0.0877936570743552\\
	241.3125	0.0701828031980669\\
	241.625	0.0514693426855764\\
	241.9375	0.0357089900523978\\
	242.25	0.0217414050702195\\
	242.5625	0.0060028622090165\\
	242.875	-0.0141492233706095\\
	243.1875	-0.0349341274272793\\
	243.5	-0.0519974435633675\\
	243.8125	-0.0600642974998785\\
	244.125	-0.0591420820320303\\
	244.4375	-0.0495943198965672\\
	244.75	-0.0358667160895775\\
	245.0625	-0.0236162154195681\\
	245.375	-0.0193326691011166\\
	245.6875	-0.0293930979426292\\
	246	-0.0526844377604389\\
	246.3125	-0.0843837213887242\\
	246.625	-0.118842177919271\\
	246.9375	-0.145555350897583\\
	247.25	-0.157261397833425\\
	247.5625	-0.14694022482058\\
	247.875	-0.111427207501132\\
	248.1875	-0.0542279299197417\\
	248.5	0.0173200358421635\\
	248.8125	0.0897972702656564\\
	249.125	0.152023788569645\\
	249.4375	0.194087484183631\\
	249.75	0.212392538813748\\
};
\addplot [color=mycolor3,solid,line width=0.8pt,forget plot]
table[row sep=crcr]{%
	125.0625	0.236631484171718\\
	125.375	0.263455128863216\\
	125.6875	0.27804991419963\\
	126	0.272857898435329\\
	126.3125	0.250239569998482\\
	126.625	0.201822132499433\\
	126.9375	0.135052119997536\\
	127.25	0.0515770358335559\\
	127.5625	-0.0380284570233571\\
	127.875	-0.123201858979113\\
	128.1875	-0.19459236937992\\
	128.5	-0.241793775410689\\
	128.8125	-0.257358273030237\\
	129.125	-0.238225179248527\\
	129.4375	-0.192423851238577\\
	129.75	-0.126249753828694\\
	130.0625	-0.0519531777796419\\
	130.375	0.0207381855867436\\
	130.6875	0.0783644856001004\\
	131	0.116754497091674\\
	131.3125	0.133196884228647\\
	131.625	0.131063023905393\\
	131.9375	0.115136993518914\\
	132.25	0.0937481738080444\\
	132.5625	0.0748484467202033\\
	132.875	0.0620067290928215\\
	133.1875	0.0579515314540589\\
	133.5	0.0591456235506568\\
	133.8125	0.06167432111733\\
	134.125	0.0621508073766448\\
	134.4375	0.0500532414569699\\
	134.75	0.0199804567124011\\
	135.0625	-0.0245887316569414\\
	135.375	-0.0806576091516086\\
	135.6875	-0.146314073640024\\
	136	-0.202193738637951\\
	136.3125	-0.244843874120472\\
	136.625	-0.262005948732913\\
	136.9375	-0.242349826313003\\
	137.25	-0.193295968084523\\
	137.5625	-0.108996315833916\\
	137.875	-0.00433946647304167\\
	138.1875	0.10719606657799\\
	138.5	0.212270063498507\\
	138.8125	0.292931173645795\\
	139.125	0.339314127063595\\
	139.4375	0.339001582774015\\
	139.75	0.291364750188916\\
	140.0625	0.199203662050739\\
	140.375	0.0748664825402856\\
	140.6875	-0.0660006955020198\\
	141	-0.20701509976888\\
	141.3125	-0.327058943257908\\
	141.625	-0.411138967498741\\
	141.9375	-0.442664880298986\\
	142.25	-0.418029314980215\\
	142.5625	-0.336482514318725\\
	142.875	-0.208034676322693\\
	143.1875	-0.0462987996080722\\
	143.5	0.129787188400969\\
	143.8125	0.296991034487682\\
	144.125	0.434078432831227\\
	144.4375	0.519349958213041\\
	144.75	0.541215450539938\\
	145.0625	0.493506683803694\\
	145.375	0.385017309653146\\
	145.6875	0.223984612043091\\
	146	0.0424359739978504\\
	146.3125	-0.137738790140004\\
	146.625	-0.291658439322883\\
	146.9375	-0.394248533729913\\
	147.25	-0.433369196008127\\
	147.5625	-0.404835699912341\\
	147.875	-0.320348553822748\\
	148.1875	-0.193297709796171\\
	148.5	-0.045756632439245\\
	148.8125	0.0968431294840203\\
	149.125	0.212109738014689\\
	149.4375	0.287750520693198\\
	149.75	0.305251449328506\\
	150.0625	0.266725773388251\\
	150.375	0.177563584544692\\
	150.6875	0.0486662564389216\\
	151	-0.101758986662995\\
	151.3125	-0.252218319751476\\
	151.625	-0.379565477348285\\
	151.9375	-0.466632082007292\\
	152.25	-0.493521948526458\\
	152.5625	-0.460774115733454\\
	152.875	-0.362318660802479\\
	153.1875	-0.208880884310209\\
	153.5	-0.0205517928316058\\
	153.8125	0.18473447954049\\
	154.125	0.376228190854943\\
	154.4375	0.535499955932059\\
	154.75	0.641775068692213\\
	155.0625	0.677899932370518\\
	155.375	0.647521931021955\\
	155.6875	0.548613894405877\\
	156	0.398222854834147\\
	156.3125	0.210017229241461\\
	156.625	0.00772601747762491\\
	156.9375	-0.185265439590023\\
	157.25	-0.353153310221151\\
	157.5625	-0.474359723005188\\
	157.875	-0.541560904066272\\
	158.1875	-0.554853491579677\\
	158.5	-0.511336295252196\\
	158.8125	-0.429932033083428\\
	159.125	-0.32083487828844\\
	159.4375	-0.201985085429151\\
	159.75	-0.088512981051706\\
	160.0625	0.00588354704754097\\
	160.375	0.0741711667248319\\
	160.6875	0.114175320162562\\
	161	0.133197080426526\\
	161.3125	0.135880726931693\\
	161.625	0.136111352418736\\
	161.9375	0.134690737997297\\
	162.25	0.141260488456612\\
	162.5625	0.155033609473318\\
	162.875	0.172773688825201\\
	163.1875	0.186687451384564\\
	163.5	0.192811957276085\\
	163.8125	0.180506089765334\\
	164.125	0.148522818883247\\
	164.4375	0.0945408343412275\\
	164.75	0.0231520376982055\\
	165.0625	-0.0575992978995238\\
	165.375	-0.138392598633328\\
	165.6875	-0.203045947011898\\
	166	-0.242943782597275\\
	166.3125	-0.253202453914104\\
	166.625	-0.223811879893573\\
	166.9375	-0.160755811422621\\
	167.25	-0.0709494126782744\\
	167.5625	0.0361963894186014\\
	167.875	0.136471640554739\\
	168.1875	0.223192271389246\\
	168.5	0.276894771074376\\
	168.8125	0.292784542467903\\
	169.125	0.266356369216471\\
	169.4375	0.203745066384089\\
	169.75	0.11218286477761\\
	170.0625	0.00533668748115886\\
	170.375	-0.0991557243794542\\
	170.6875	-0.187578982982591\\
	171	-0.247264294679272\\
	171.3125	-0.277683739594105\\
	171.625	-0.271478378610257\\
	171.9375	-0.236242704380679\\
	172.25	-0.179151394979614\\
	172.5625	-0.1107484540924\\
	172.875	-0.0395870534529434\\
	173.1875	0.0276411379731342\\
	173.5	0.0800406511497581\\
	173.8125	0.118744336879495\\
	174.125	0.138263852507589\\
	174.4375	0.145563290215041\\
	174.75	0.138549175627064\\
	175.0625	0.121721823402706\\
	175.375	0.0990026358138615\\
	175.6875	0.0718029799177057\\
	176	0.0427864592045767\\
	176.3125	0.0153894013603543\\
	176.625	-0.0106876339093329\\
	176.9375	-0.033586436019105\\
	177.25	-0.0480709692054683\\
	177.5625	-0.0629083797696425\\
	177.875	-0.0697318053616231\\
	178.1875	-0.0738349309243297\\
	178.5	-0.0750699472212503\\
	178.8125	-0.0741547986780935\\
	179.125	-0.0686212475655957\\
	179.4375	-0.0658988712959807\\
	179.75	-0.0601665976494142\\
	180.0625	-0.0522717597571887\\
	180.375	-0.0427446706431896\\
	180.6875	-0.0342398241757116\\
	181	-0.020244219940295\\
	181.3125	-0.00600101669532318\\
	181.625	0.0109393345008271\\
	181.9375	0.031856218287156\\
	182.25	0.0558307939568196\\
	182.5625	0.0812257864960278\\
	182.875	0.110652440466419\\
	183.1875	0.135868792265829\\
	183.5	0.16000821021137\\
	183.8125	0.177042387009282\\
	184.125	0.18172103228275\\
	184.4375	0.168731982267269\\
	184.75	0.137034016529467\\
	185.0625	0.0877314193365113\\
	185.375	0.0229088276682928\\
	185.6875	-0.0505021787506926\\
	186	-0.124725588622821\\
	186.3125	-0.185557982143975\\
	186.625	-0.233078429639373\\
	186.9375	-0.252697498948272\\
	187.25	-0.241864095870546\\
	187.5625	-0.204138181571578\\
	187.875	-0.140834476532738\\
	188.1875	-0.0601340736377383\\
	188.5	0.0266512162044161\\
	188.8125	0.107944114635287\\
	189.125	0.173833973417691\\
	189.4375	0.212485585018261\\
	189.75	0.22254449878749\\
	190.0625	0.198023992460008\\
	190.375	0.148080377951205\\
	190.6875	0.0753368613703753\\
	191	-0.0077407988665684\\
	191.3125	-0.0898919056593856\\
	191.625	-0.157231187835261\\
	191.9375	-0.203875744398279\\
	192.25	-0.227143315210344\\
	192.5625	-0.220376643542439\\
	192.875	-0.191535932050874\\
	193.1875	-0.139598444834669\\
	193.5	-0.0739083098120268\\
	193.8125	0.0040902624064283\\
	194.125	0.0833032463157053\\
	194.4375	0.161427362608904\\
	194.75	0.230095826390543\\
	195.0625	0.282007536578803\\
	195.375	0.317496207935179\\
	195.6875	0.318060127711124\\
	196	0.289987904702516\\
	196.3125	0.227428000439241\\
	196.625	0.135326293120631\\
	196.9375	0.0171372340223032\\
	197.25	-0.115288587317846\\
	197.5625	-0.245644301560769\\
	197.875	-0.361065148891059\\
	198.1875	-0.447071154477911\\
	198.5	-0.495227287723769\\
	198.8125	-0.49330544782791\\
	199.125	-0.443212391339978\\
	199.4375	-0.349573261590504\\
	199.75	-0.219727800777464\\
	200.0625	-0.0679828727750168\\
	200.375	0.0890623036813992\\
	200.6875	0.241674164133204\\
	201	0.370081896131027\\
	201.3125	0.466522073314295\\
	201.625	0.521117267646495\\
	201.9375	0.535240186655018\\
	202.25	0.506776237184011\\
	202.5625	0.440419451886631\\
	202.875	0.343109666637781\\
	203.1875	0.220913840873099\\
	203.5	0.0788044847428765\\
	203.8125	-0.0698017452020316\\
	204.125	-0.218638344658923\\
	204.4375	-0.355786287281465\\
	204.75	-0.469188140480503\\
	205.0625	-0.548269442380839\\
	205.375	-0.579654529457303\\
	205.6875	-0.557191029437142\\
	206	-0.483268598088795\\
	206.3125	-0.357906965077465\\
	206.625	-0.192416791895996\\
	206.9375	-0.00774985090666384\\
	207.25	0.178375783406381\\
	207.5625	0.341698070354214\\
	207.875	0.463539323960571\\
	208.1875	0.530807769842629\\
	208.5	0.532219064809963\\
	208.8125	0.468818159085157\\
	209.125	0.356009856355641\\
	209.4375	0.208627845758718\\
	209.75	0.0479247629074742\\
	210.0625	-0.101063184635363\\
	210.375	-0.217319235137257\\
	210.6875	-0.286797438971324\\
	211	-0.296819569318079\\
	211.3125	-0.250113119711813\\
	211.625	-0.15653934879515\\
	211.9375	-0.0339963218007483\\
	212.25	0.0986130646816172\\
	212.5625	0.213387455512194\\
	212.875	0.293143484149395\\
	213.1875	0.320529606540608\\
	213.5	0.291261960285125\\
	213.8125	0.205605958242979\\
	214.125	0.0773532213452882\\
	214.4375	-0.0700159915568687\\
	214.75	-0.219205876200004\\
	215.0625	-0.342256582508521\\
	215.375	-0.423979672599052\\
	215.6875	-0.451054201721334\\
	216	-0.419581211187737\\
	216.3125	-0.333939975907513\\
	216.625	-0.211050493937282\\
	216.9375	-0.0637069534159105\\
	217.25	0.0858027658842447\\
	217.5625	0.21997113527304\\
	217.875	0.321998939356957\\
	218.1875	0.383194423111\\
	218.5	0.395882253860797\\
	218.8125	0.364674255786998\\
	219.125	0.291796053286397\\
	219.4375	0.187126594366053\\
	219.75	0.0620023159229448\\
	220.0625	-0.0643904815116229\\
	220.375	-0.185091376855624\\
	220.6875	-0.287463838875382\\
	221	-0.358172322576502\\
	221.3125	-0.387043609157888\\
	221.625	-0.372186795809238\\
	221.9375	-0.315035440200092\\
	222.25	-0.218123908215915\\
	222.5625	-0.0919830010003745\\
	222.875	0.0514000838381151\\
	223.1875	0.190880272021222\\
	223.5	0.314721852557441\\
	223.8125	0.398860208727419\\
	224.125	0.435983796099864\\
	224.4375	0.415154521145867\\
	224.75	0.340808189430442\\
	225.0625	0.217370317615434\\
	225.375	0.0570821379184621\\
	225.6875	-0.113642292062832\\
	226	-0.272445670404911\\
	226.3125	-0.401519717401707\\
	226.625	-0.476839528414577\\
	226.9375	-0.489991752400935\\
	227.25	-0.439522172307991\\
	227.5625	-0.330876997154197\\
	227.875	-0.178501880989446\\
	228.1875	-0.00848373927858014\\
	228.5	0.159530762811669\\
	228.8125	0.301974386948287\\
	229.125	0.397450031401855\\
	229.4375	0.439176558395637\\
	229.75	0.421043596326879\\
	230.0625	0.35455448354631\\
	230.375	0.24491372751394\\
	230.6875	0.113125746261048\\
	231	-0.0230546114293276\\
	231.3125	-0.144864697880425\\
	231.625	-0.240248182827299\\
	231.9375	-0.29731668789913\\
	232.25	-0.313179989153268\\
	232.5625	-0.291751905659318\\
	232.875	-0.235162605498931\\
	233.1875	-0.154547571797249\\
	233.5	-0.0622971999822905\\
	233.8125	0.0279611972075236\\
	234.125	0.110344332432457\\
	234.4375	0.173602842888666\\
	234.75	0.217248983276218\\
	235.0625	0.235229204360727\\
	235.375	0.227316071673986\\
	235.6875	0.196680428974414\\
	236	0.149096981867794\\
	236.3125	0.0867446923395481\\
	236.625	0.0185213835338357\\
	236.9375	-0.0492773118703478\\
	237.25	-0.113474048774324\\
	237.5625	-0.159822239102812\\
	237.875	-0.192907348797036\\
	238.1875	-0.204351082058743\\
	238.5	-0.192325599271526\\
	238.8125	-0.159261626791704\\
	239.125	-0.112053980491438\\
	239.4375	-0.051555574289414\\
	239.75	0.0113421344189935\\
	240.0625	0.0650527832125472\\
	240.375	0.106972318002275\\
	240.6875	0.132796623922443\\
	241	0.140395083502009\\
	241.3125	0.132094212202449\\
	241.625	0.10857797085608\\
	241.9375	0.0779254945047997\\
	242.25	0.04527383521745\\
	242.5625	0.0150354237437198\\
	242.875	-0.0115650846039642\\
	243.1875	-0.0268088163310474\\
	243.5	-0.033298228235478\\
	243.8125	-0.0285831230194521\\
	244.125	-0.0203551851746689\\
	244.4375	-0.0101790204126411\\
	244.75	-0.00553854235498638\\
	245.0625	-0.0075391859102107\\
	245.375	-0.0160279339383038\\
	245.6875	-0.0372668514398238\\
	246	-0.0659958759748537\\
	246.3125	-0.0969811862727079\\
	246.625	-0.128553376006\\
	246.9375	-0.146846824036292\\
	247.25	-0.150467885170835\\
	247.5625	-0.135207441816008\\
	247.875	-0.0998318846770054\\
	248.1875	-0.0497049965401641\\
	248.5	0.012940801665482\\
	248.8125	0.074749976007447\\
	249.125	0.130501784796384\\
	249.4375	0.171585609986932\\
	249.75	0.194050410346352\\
};
\addplot [color=mycolor4,solid,line width=0.8pt,forget plot]
table[row sep=crcr]{%
	125.0625	0.0545439335981682\\
	125.375	0.0648638984553637\\
	125.6875	0.0723302651567415\\
	126	0.0759420876156693\\
	126.3125	0.0764058902697801\\
	126.625	0.0709484463445355\\
	126.9375	0.0582716530324316\\
	127.25	0.0372564423192535\\
	127.5625	0.0113036958148413\\
	127.875	-0.0152794623923914\\
	128.1875	-0.0384643800673445\\
	128.5	-0.0555027313886621\\
	128.8125	-0.0638093792812379\\
	129.125	-0.0609545211592388\\
	129.4375	-0.0472808246645158\\
	129.75	-0.02493209168659\\
	130.0625	0.000401488758644814\\
	130.375	0.02223116956299\\
	130.6875	0.0352716164129541\\
	131	0.0389862460810443\\
	131.3125	0.0359149366872078\\
	131.625	0.0300945772652737\\
	131.9375	0.0241171757228262\\
	132.25	0.0191901721657678\\
	132.5625	0.0150571892081925\\
	132.875	0.0120503537514836\\
	133.1875	0.0102013259748013\\
	133.5	0.0104978304210663\\
	133.8125	0.0126341704748584\\
	134.125	0.0160721856061571\\
	134.4375	0.0177065695928289\\
	134.75	0.014564214809816\\
	135.0625	0.00436062147825957\\
	135.375	-0.0132138074725884\\
	135.6875	-0.0368337437126086\\
	136	-0.0601327041183088\\
	136.3125	-0.0787249902800238\\
	136.625	-0.0877422534577378\\
	136.9375	-0.0851843457739536\\
	137.25	-0.0720646917067601\\
	137.5625	-0.0476774138023751\\
	137.875	-0.0149953619877147\\
	138.1875	0.0205453225157092\\
	138.5	0.0525102997827945\\
	138.8125	0.0752405325186707\\
	139.125	0.0870312066729978\\
	139.4375	0.0875179393712269\\
	139.75	0.0768548523022994\\
	140.0625	0.0545149196846267\\
	140.375	0.0214042293623849\\
	140.6875	-0.0184305655770434\\
	141	-0.0588291226456438\\
	141.3125	-0.0916450091219475\\
	141.625	-0.111201545980948\\
	141.9375	-0.114048704000858\\
	142.25	-0.102056473156044\\
	142.5625	-0.0791814909149593\\
	142.875	-0.0494473342588063\\
	143.1875	-0.0140672430023671\\
	143.5	0.027379232054881\\
	143.8125	0.0727228660099481\\
	144.125	0.116648950316508\\
	144.4375	0.1506855452063\\
	144.75	0.167830299692852\\
	145.0625	0.162395922090585\\
	145.375	0.133990435915704\\
	145.6875	0.0837980569504134\\
	146	0.0206049294381837\\
	146.3125	-0.0449973679277914\\
	146.625	-0.100180288016607\\
	146.9375	-0.133585602226087\\
	147.25	-0.140571004963146\\
	147.5625	-0.122862465624369\\
	147.875	-0.0899503246725321\\
	148.1875	-0.0506751692333072\\
	148.5	-0.0138340492066883\\
	148.8125	0.0155590163752133\\
	149.125	0.0361552723468994\\
	149.4375	0.0489672479658886\\
	149.75	0.0531925704134498\\
	150.0625	0.0481878626987003\\
	150.375	0.0325542168201908\\
	150.6875	0.00666818209192799\\
	151	-0.0263523594909126\\
	151.3125	-0.0615532108971465\\
	151.625	-0.0927275567926216\\
	151.9375	-0.113907365285538\\
	152.25	-0.11905202490704\\
	152.5625	-0.107573496059857\\
	152.875	-0.07951791981472\\
	153.1875	-0.0403354538000831\\
	153.5	0.00372718294023555\\
	153.8125	0.0477204510548252\\
	154.125	0.0858168918954343\\
	154.4375	0.114488114266082\\
	154.75	0.130706387343118\\
	155.0625	0.132070096377203\\
	155.375	0.121831335170366\\
	155.6875	0.100823675578231\\
	156	0.0730613171809803\\
	156.3125	0.0404482603128847\\
	156.625	0.0061083283541241\\
	156.9375	-0.0273679634217294\\
	157.25	-0.0579476883728296\\
	157.5625	-0.0820353351383033\\
	157.875	-0.0967514307922739\\
	158.1875	-0.100456174695461\\
	158.5	-0.0912090028165248\\
	158.8125	-0.0741887142981663\\
	159.125	-0.0531341520240791\\
	159.4375	-0.0328697944392743\\
	159.75	-0.0155288621084915\\
	160.0625	-0.00217442628321611\\
	160.375	0.00619046991581008\\
	160.6875	0.00922523280388392\\
	161	0.00810359695392561\\
	161.3125	0.00566206966659545\\
	161.625	0.00510632858997932\\
	161.9375	0.00708072427038994\\
	162.25	0.0113550014503035\\
	162.5625	0.0167267258431868\\
	162.875	0.0220808144693162\\
	163.1875	0.0261232229341116\\
	163.5	0.0282495656359352\\
	163.8125	0.0272125146759604\\
	164.125	0.0233540032716937\\
	164.4375	0.0171761234612771\\
	164.75	0.00948370943973694\\
	165.0625	0.00115305959103291\\
	165.375	-0.00720649783422926\\
	165.6875	-0.0143274472427332\\
	166	-0.0198245241189322\\
	166.3125	-0.0229840023874358\\
	166.625	-0.0219682729403584\\
	166.9375	-0.0165308480327825\\
	167.25	-0.0070367941177504\\
	167.5625	0.00512271484796458\\
	167.875	0.0167726236659967\\
	168.1875	0.0267874386651961\\
	168.5	0.0328703486836037\\
	168.8125	0.0342191559700731\\
	169.125	0.0299318659134859\\
	169.4375	0.0205828088716189\\
	169.75	0.00708808624665075\\
	170.0625	-0.00785668892093062\\
	170.375	-0.0213323788141041\\
	170.6875	-0.0310397937056891\\
	171	-0.0358503567253288\\
	171.3125	-0.0366874441540762\\
	171.625	-0.0333186017792081\\
	171.9375	-0.0271468551459102\\
	172.25	-0.0191549460703966\\
	172.5625	-0.0109338053537713\\
	172.875	-0.003884237501536\\
	173.1875	0.00106866043069635\\
	173.5	0.0039047871170967\\
	173.8125	0.00605085574227377\\
	174.125	0.00771653465605508\\
	174.4375	0.00908359368049117\\
	174.75	0.00891201384989956\\
	175.0625	0.00737923531904044\\
	175.375	0.00592686010363933\\
	175.6875	0.00554622011303373\\
	176	0.00580486924349629\\
	176.3125	0.00521403672221156\\
	176.625	0.00215802416498336\\
	176.9375	-0.00314851146908177\\
	177.25	-0.00851873227275151\\
	177.5625	-0.0129857026455332\\
	177.875	-0.0143618384761133\\
	178.1875	-0.0132467956646546\\
	178.5	-0.0105560537788009\\
	178.8125	-0.00779416363542394\\
	179.125	-0.00582232798723939\\
	179.4375	-0.0056930215250702\\
	179.75	-0.00640560901609577\\
	180.0625	-0.00727005437633758\\
	180.375	-0.00734394784787513\\
	180.6875	-0.00626652140280657\\
	181	-0.00296167930821277\\
	181.3125	0.00189533770639199\\
	181.625	0.00781967861695923\\
	181.9375	0.0134637463182959\\
	182.25	0.0170817084129935\\
	182.5625	0.0178650950053064\\
	182.875	0.017239442347751\\
	183.1875	0.0162461524415207\\
	183.5	0.0175215972400385\\
	183.8125	0.0212553704943997\\
	184.125	0.0257985000825887\\
	184.4375	0.0283155148156791\\
	184.75	0.0266550554110357\\
	185.0625	0.0197718689870643\\
	185.375	0.00876688319811041\\
	185.6875	-0.00389463685221666\\
	186	-0.0151972387345793\\
	186.3125	-0.0226543813987444\\
	186.625	-0.0269289918971398\\
	186.9375	-0.0274101612502509\\
	187.25	-0.0253118216771502\\
	187.5625	-0.0221307411779863\\
	187.875	-0.017764668045216\\
	188.1875	-0.0118485155791279\\
	188.5	-0.00378970265769378\\
	188.8125	0.00610391163929865\\
	189.125	0.0161642436382149\\
	189.4375	0.0235015499176055\\
	189.75	0.0267773661303942\\
	190.0625	0.0248564087720805\\
	190.375	0.0187265062324187\\
	190.6875	0.00812403571466975\\
	191	-0.0055614891606235\\
	191.3125	-0.01971441912023\\
	191.625	-0.0303219763332385\\
	191.9375	-0.0343654350714811\\
	192.25	-0.0318663214588724\\
	192.5625	-0.0240052374014137\\
	192.875	-0.0154617127216871\\
	193.1875	-0.00813200919684979\\
	193.5	-0.00302347521413531\\
	193.8125	0.00142823137105601\\
	194.125	0.00637725644620514\\
	194.4375	0.0130810387164346\\
	194.75	0.0210790662282098\\
	195.0625	0.0291362772380826\\
	195.375	0.0368629374086599\\
	195.6875	0.0403047279095032\\
	196	0.0399576592273159\\
	196.3125	0.0337462388460935\\
	196.625	0.0219843360148444\\
	196.9375	0.00471755132910671\\
	197.25	-0.0166487750818569\\
	197.5625	-0.0402573388527444\\
	197.875	-0.0635365279641071\\
	198.1875	-0.0823169434143137\\
	198.5	-0.0932582336969636\\
	198.8125	-0.0919455600423253\\
	199.125	-0.0801186030927667\\
	199.4375	-0.0610472241953966\\
	199.75	-0.0373690794338654\\
	200.0625	-0.0118078509947927\\
	200.375	0.0141364861926988\\
	200.6875	0.0398723335646853\\
	201	0.0628068366137993\\
	201.3125	0.0814145748254486\\
	201.625	0.0923787621003485\\
	201.9375	0.0956157277464165\\
	202.25	0.0896400204280861\\
	202.5625	0.0763391998825277\\
	202.875	0.0587327313079061\\
	203.1875	0.0396428994728622\\
	203.5	0.019521196971671\\
	203.8125	-0.00195716705935579\\
	204.125	-0.0272893030999582\\
	204.4375	-0.0559784710770164\\
	204.75	-0.0844548632368933\\
	205.0625	-0.107506100169526\\
	205.375	-0.119911371210435\\
	205.6875	-0.119524846885883\\
	206	-0.107878152429104\\
	206.3125	-0.0836475775736671\\
	206.625	-0.0479651985648676\\
	206.9375	-0.00399957959905232\\
	207.25	0.0429539122393969\\
	207.5625	0.0852976641344718\\
	207.875	0.116693234373714\\
	208.1875	0.134113867304851\\
	208.5	0.135191791732651\\
	208.8125	0.120503492343759\\
	209.125	0.0945700102379212\\
	209.4375	0.0606757334951939\\
	209.75	0.0245260532723651\\
	210.0625	-0.00818698156694863\\
	210.375	-0.0335298585312814\\
	210.6875	-0.0500362168958126\\
	211	-0.0562399771563931\\
	211.3125	-0.0517006612166092\\
	211.625	-0.0353554862744591\\
	211.9375	-0.00807125067852218\\
	212.25	0.0268298392334452\\
	212.5625	0.0610823674886268\\
	212.875	0.0862999995305788\\
	213.1875	0.0944959917178887\\
	213.5	0.083498911537227\\
	213.8125	0.0543519852332066\\
	214.125	0.0129350368752441\\
	214.4375	-0.0332027714697684\\
	214.75	-0.0776549291775958\\
	215.0625	-0.113100301705979\\
	215.375	-0.134871734299933\\
	215.6875	-0.139355496422186\\
	216	-0.126194021787063\\
	216.3125	-0.0981921630451302\\
	216.625	-0.0613714670535156\\
	216.9375	-0.0198519990866664\\
	217.25	0.0203123889204901\\
	217.5625	0.0544471739964711\\
	217.875	0.0781710111713631\\
	218.1875	0.0903423203651099\\
	218.5	0.0911747427111622\\
	218.8125	0.0827047037555173\\
	219.125	0.0654574816143465\\
	219.4375	0.0403967080817922\\
	219.75	0.0100047526027293\\
	220.0625	-0.0216268401148722\\
	220.375	-0.0518724807707739\\
	220.6875	-0.0793610686722609\\
	221	-0.101916763878368\\
	221.3125	-0.115944533492263\\
	221.625	-0.117899818830662\\
	221.9375	-0.105491758802548\\
	222.25	-0.0782718568542083\\
	222.5625	-0.0400229200299782\\
	222.875	0.00378150517306452\\
	223.1875	0.0469360236873301\\
	223.5	0.0849252289112804\\
	223.8125	0.112603794303851\\
	224.125	0.127401602153389\\
	224.4375	0.125917089325697\\
	224.75	0.107217550422179\\
	225.0625	0.0724656884742824\\
	225.375	0.0249497396588774\\
	225.6875	-0.0287940502602844\\
	226	-0.0819074603199204\\
	226.3125	-0.127993715880178\\
	226.625	-0.15798711102076\\
	226.9375	-0.166407930105502\\
	227.25	-0.150815789413317\\
	227.5625	-0.113164491220775\\
	227.875	-0.0607371579380449\\
	228.1875	-0.00374036675838609\\
	228.5	0.0491709753456889\\
	228.8125	0.0915286449884069\\
	229.125	0.117710808739613\\
	229.4375	0.125806847720226\\
	229.75	0.115124008982949\\
	230.0625	0.0910388476100687\\
	230.375	0.0580972988809033\\
	230.6875	0.0235140841380668\\
	231	-0.00884743484803397\\
	231.3125	-0.0361504143468724\\
	231.625	-0.0559594978247019\\
	231.9375	-0.0659998646385245\\
	232.25	-0.065624852023847\\
	232.5625	-0.0565275704369006\\
	232.875	-0.0407416195794514\\
	233.1875	-0.0211477707412902\\
	233.5	-9.38118431480949e-05\\
	233.8125	0.0197848717265286\\
	234.125	0.0359645615362618\\
	234.4375	0.045310275405173\\
	234.75	0.0473731154104869\\
	235.0625	0.0431355007466297\\
	235.375	0.0358801010368965\\
	235.6875	0.0284975546114415\\
	236	0.0218780791676846\\
	236.3125	0.0141273724544417\\
	236.625	0.00508092610186111\\
	236.9375	-0.00464918862914003\\
	237.25	-0.013643316346721\\
	237.5625	-0.0204646950580575\\
	237.875	-0.0259737307133067\\
	238.1875	-0.029508650975584\\
	238.5	-0.0301699901241444\\
	238.8125	-0.026934907908145\\
	239.125	-0.0197510107846828\\
	239.4375	-0.00866302169597048\\
	239.75	0.0033515908427784\\
	240.0625	0.0126328727183836\\
	240.375	0.0169654838266873\\
	240.6875	0.0161145429767485\\
	241	0.0122938993104434\\
	241.3125	0.00850857172465742\\
	241.625	0.00609429180037071\\
	241.9375	0.00532729002506298\\
	242.25	0.00460606779233296\\
	242.5625	0.00261205485320255\\
	242.875	-0.00123615302311181\\
	243.1875	-0.00520215619639413\\
	243.5	-0.00820612686077637\\
	243.8125	-0.00900723961737371\\
	244.125	-0.00835074316281846\\
	244.4375	-0.00644263748641988\\
	244.75	-0.00429311987490292\\
	245.0625	-0.00238785706829634\\
	245.375	-0.00156808557096809\\
	245.6875	-0.0030248617511465\\
	246	-0.00623881488579626\\
	246.3125	-0.0100545405650806\\
	246.625	-0.0140626894571637\\
	246.9375	-0.0167931287006431\\
	247.25	-0.0183346214627934\\
	247.5625	-0.0177954222069227\\
	247.875	-0.0143088377664883\\
	248.1875	-0.0078397236416963\\
	248.5	0.00102669819480227\\
	248.8125	0.00972194625171421\\
	249.125	0.016395771510048\\
	249.4375	0.0197466766798575\\
	249.75	0.0202956536444391\\
};
\end{axis}
\end{tikzpicture}%

%% file: figs/noise_reduction_comparison.tex
%
%
\definecolor{mycolor1}{rgb}{0.26667,0.44706,0.70980}%
\definecolor{mycolor2}{rgb}{0.16471,0.67059,0.38039}%
\definecolor{mycolor3}{rgb}{0.82745,0.26275,0.30588}%
\begin{tikzpicture}

\begin{axis}[%
xmin=0.01,
xmax=0.5,
xtick={0.01,  0.1,  0.3,  0.5},
xlabel style={font=\color{white!15!black}},
xlabel={Corrupting noise variance},
ymin=0,
ymax=15,
ylabel style={font=\color{white!15!black}},
ylabel={SNR [dB]},
axis background/.style={fill=white},
legend style={legend cell align=left, align=left, draw=white!15!black,fill=white},
width=\figurewidth,
height=\figureheight
]
\addplot [color=mycolor1, dashed, line width=1.0pt]
  table[row sep=crcr]{%
0.01	4.44857381766576\\
0.05	3.54502316964845\\
0.1	2.87349524552674\\
0.3	1.67190882335775\\
0.5	1.18556833318509\\
};
\addlegendentry{EP 1}

\addplot [color=mycolor1, line width=1.0pt]
  table[row sep=crcr]{%
0.01	9.4990363444523\\
0.05	9.24973680511635\\
0.1	7.97652192318652\\
0.3	5.41127386147334\\
0.5	4.22531580303243\\
};
\addlegendentry{EP 20}

\addplot [color=mycolor2, dashed, line width=1.0pt]
  table[row sep=crcr]{%
0.01	5.27913108976448\\
0.05	3.9068407053838\\
0.1	2.96889320382806\\
0.3	1.72110889874384\\
0.5	1.17992857254459\\
};
\addlegendentry{IHGP 1}

\addplot [color=mycolor2, line width=1.0pt]
  table[row sep=crcr]{%
0.01	8.97170922494787\\
0.05	6.18157874439819\\
0.1	4.42618721933355\\
0.3	2.5217688678236\\
0.5	1.75895560929121\\
};
\addlegendentry{IHGP 20}

\addplot [color=mycolor3, dashed, line width=1.0pt]
  table[row sep=crcr]{%
0.01	5.4730488252878\\
0.05	5.50392577634777\\
0.1	5.3506259096377\\
0.3	4.90748046720286\\
0.5	4.60206638671285\\
};
\addlegendentry{EKF 1}

\addplot [color=mycolor3, line width=1.0pt]
  table[row sep=crcr]{%
0.01	5.4474733503891\\
0.05	5.46986423865696\\
0.1	5.37373138061364\\
0.3	4.92645469366407\\
0.5	4.62610688194024\\
};
\addlegendentry{EKF 20}

\addplot [color=gray, line width=1.0pt]
  table[row sep=crcr]{%
0.01	13.02996553111\\
0.05	10.6289668226327\\
0.1	9.41036810831252\\
0.3	7.39149284594851\\
0.5	6.4164170945103\\
};
\addlegendentry{SpecSub}

\addplot[area legend, draw=black, fill=mycolor1, draw opacity=0, fill opacity=0.15, forget plot]
table[row sep=crcr] {%
x	y\\
0.01	3.2218091627859\\
0.05	2.29110960382811\\
0.1	1.67939693729972\\
0.3	0.829633039543561\\
0.5	0.535117253387528\\
0.5	1.83601941298265\\
0.3	2.51418460717193\\
0.1	4.06759355375375\\
0.05	4.7989367354688\\
0.01	5.67533847254563\\
}--cycle;

\addplot[area legend, draw=black, fill=mycolor1, draw opacity=0, fill opacity=0.15, forget plot]
table[row sep=crcr] {%
x	y\\
0.01	6.88739691040767\\
0.05	7.00023064109271\\
0.1	6.20445162469474\\
0.3	4.09337454075091\\
0.5	3.13661850024172\\
0.5	5.31401310582314\\
0.3	6.72917318219577\\
0.1	9.7485922216783\\
0.05	11.49924296914\\
0.01	12.1106757784969\\
}--cycle;

\addplot[area legend, draw=black, fill=mycolor2, draw opacity=0, fill opacity=0.15, forget plot]
table[row sep=crcr] {%
x	y\\
0.01	3.95459912873118\\
0.05	2.62246287792152\\
0.1	1.68803999340403\\
0.3	0.798521371456888\\
0.5	0.540625579851743\\
0.5	1.81923156523745\\
0.3	2.64369642603079\\
0.1	4.24974641425208\\
0.05	5.19121853284608\\
0.01	6.60366305079778\\
}--cycle;

\addplot[area legend, draw=black, fill=mycolor2, draw opacity=0, fill opacity=0.15, forget plot]
table[row sep=crcr] {%
x	y\\
0.01	7.2680014221785\\
0.05	5.08344019196155\\
0.1	3.22862507545993\\
0.3	1.56861277350774\\
0.5	0.93601561003253\\
0.5	2.58189560854989\\
0.3	3.47492496213945\\
0.1	5.62374936320716\\
0.05	7.27971729683482\\
0.01	10.6754170277172\\
}--cycle;

\addplot[area legend, draw=black, fill=mycolor3, draw opacity=0, fill opacity=0.15, forget plot]
table[row sep=crcr] {%
x	y\\
0.01	3.98986561379107\\
0.05	4.14801696596767\\
0.1	4.10968946064298\\
0.3	3.7819511382785\\
0.5	3.63575855906456\\
0.5	5.56837421436114\\
0.3	6.03300979612722\\
0.1	6.59156235863241\\
0.05	6.85983458672787\\
0.01	6.95623203678453\\
}--cycle;

\addplot[area legend, draw=black, fill=mycolor3, draw opacity=0, fill opacity=0.15, forget plot]
table[row sep=crcr] {%
x	y\\
0.01	3.98039145691138\\
0.05	4.14080314748681\\
0.1	4.12624637581717\\
0.3	3.8275872145701\\
0.5	3.65312303723622\\
0.5	5.59909072664426\\
0.3	6.02532217275805\\
0.1	6.6212163854101\\
0.05	6.79892532982711\\
0.01	6.91455524386682\\
}--cycle;

\addplot[area legend, draw=black, fill=gray, draw opacity=0, fill opacity=0.15, forget plot]
table[row sep=crcr] {%
x	y\\
0.01	12.3921481148768\\
0.05	9.74994582887927\\
0.1	8.4785259908335\\
0.3	6.43217442242704\\
0.5	5.44376181683426\\
0.5	7.38907237218633\\
0.3	8.35081126946998\\
0.1	10.3422102257915\\
0.05	11.5079878163861\\
0.01	13.6677829473431\\
}--cycle;
\end{axis}
\end{tikzpicture}%

%% file: figs/noise_reduction_clean.tex
%
%
\begin{tikzpicture}

\begin{axis}[%
axis on top,
xmin=-0.5,
xmax=1455.5,
xlabel style={font=\color{white!15!black}},
ymin=0.5,
ymax=16.5,
ylabel style={font=\color{white!15!black}},
ylabel={Frequency channel \#},
ytick={4.5,8.5,12.5},
yticklabels={},
xticklabels={},
axis background/.style={fill=white},
legend style={legend cell align=left, align=left, draw=white!15!black},
width=\figurewidth,
height=\figureheight
]
\addplot [forget plot] graphics [xmin=-0.5, xmax=1455.5, ymin=0.5, ymax=16.5] {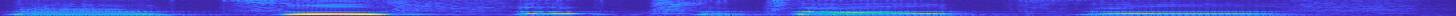};
\end{axis}
\end{tikzpicture}%

%% file: figs/noise_reduction_noisey.tex
%
%
\begin{tikzpicture}

\begin{axis}[%
axis on top,
xmin=-0.5,
xmax=1455.5,
xlabel style={font=\color{white!15!black}},
ymin=0.5,
ymax=16.5,
ylabel style={font=\color{white!15!black}},
ylabel={Frequency channel \#},
ytick={4.5,8.5,12.5},
yticklabels={},
xticklabels={},
axis background/.style={fill=white},
legend style={legend cell align=left, align=left, draw=white!15!black},
width=\figurewidth,
height=\figureheight
]
\addplot [forget plot] graphics [xmin=0.5, xmax=1456.5, ymin=0.5, ymax=16.5] {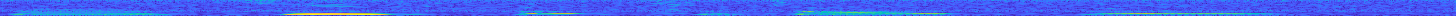};
\end{axis}
\end{tikzpicture}%

%% file: figs/noise_reduction_recon.tex
%
%
\begin{tikzpicture}

\begin{axis}[%
axis on top,
xmin=-0.5,
xmax=1455.5,
xlabel style={font=\color{white!15!black}},
xlabel={Time [ms]},
ymin=0.5,
ymax=16.5,
ylabel style={font=\color{white!15!black}},
ylabel={Frequency channel \#},
ytick={4.5,8.5,12.5},
yticklabels={},
axis background/.style={fill=white},
legend style={legend cell align=left, align=left, draw=white!15!black},
width=\figurewidth,
height=\figureheight
]
\addplot [forget plot] graphics [xmin=0.5, xmax=1456.5, ymin=0.5, ymax=16.5] {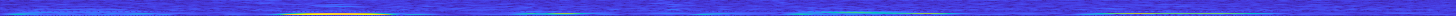};
\end{axis}
\end{tikzpicture}%

%% file: figs/source_sep.tex
%
%
\begin{tikzpicture}

\begin{axis}[%
axis on top,
xmin=1,
xmax=6,
xlabel={Time [secs]},
y dir=reverse,
ymin=-23,
ymax=16.5,
ytick={\empty},
axis background/.style={fill=white},
legend style={legend cell align=left,align=left,draw=white!15!black},
width=\figurewidth,
height=\figureheight
]
\addplot [forget plot] graphics [xmin=0.998301630434783,xmax=6.00169836956522,ymin=-23.0228060046189,ymax=16.5228060046189] {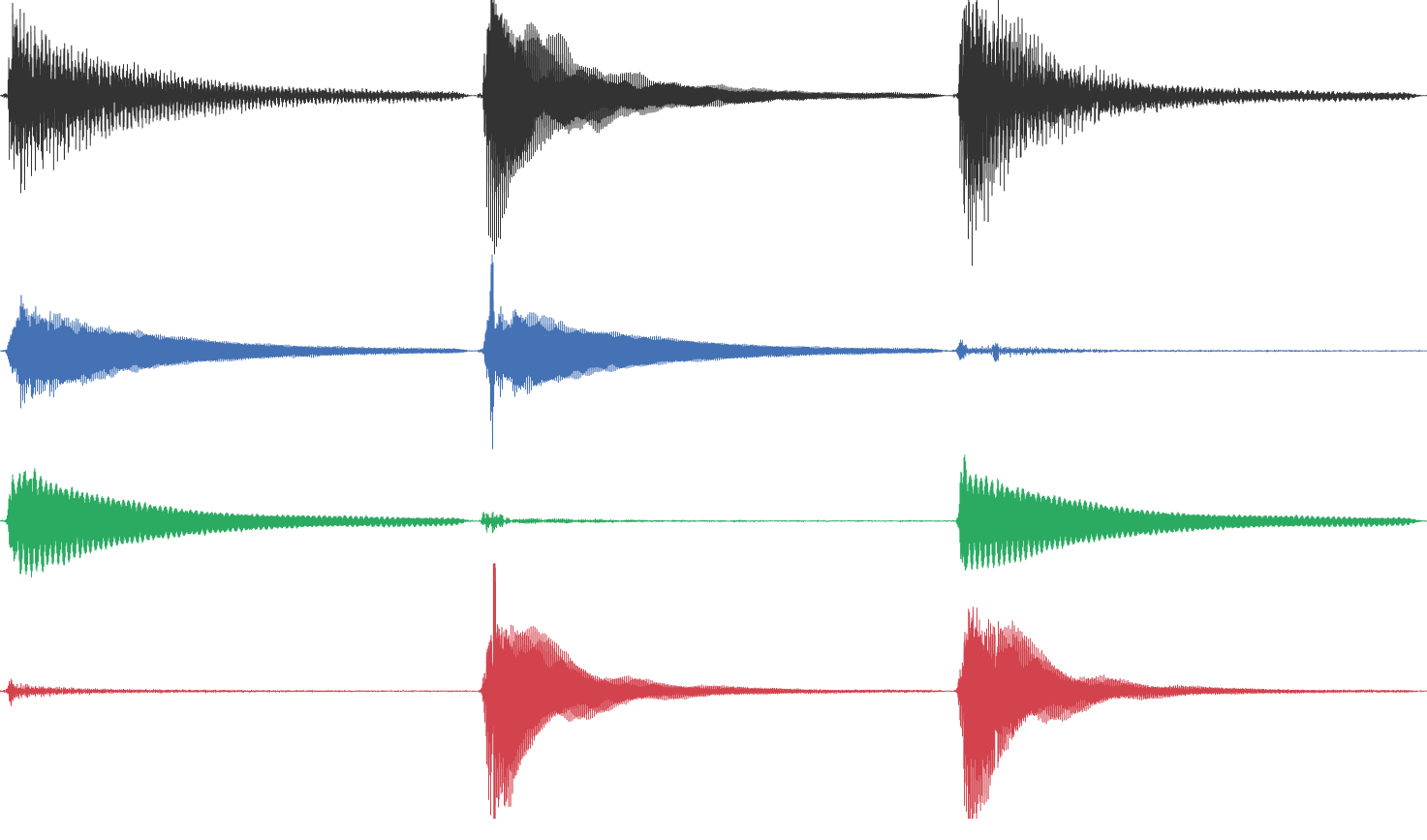};
\node[left, align=right, text=black]
at (axis cs:6,-22) {\tiny\bf Input audio, $y$};
\node[left, align=right, text=black]
at (axis cs:6,-10) {\tiny\textcolor{mycolor0}{\bf Source one: piano note C}};
\node[left, align=right, text=black]
at (axis cs:6,-2) {\tiny\textcolor{mycolor1}{\bf Source two: piano note E}};
\node[left, align=right, text=black]
at (axis cs:6,6) {\tiny\textcolor{mycolor2}{\bf Source three: piano note G}};
\end{axis}
\end{tikzpicture}%